\documentclass[10pt,twocolumn,letterpaper]{article}

\usepackage[pagenumbers]{cvpr} 

\usepackage[
    backend=biber,
    style=apa,
    sorting=nyt
]{biblatex}

\usepackage{float}
\usepackage[section]{placeins}

\usepackage{tikz}
\usetikzlibrary{positioning,arrows.meta}

\usepackage{pgfplots}
\pgfplotsset{compat=1.18}

\usepackage{dblfloatfix}

\usepackage{cuted}

\definecolor{cvprblue}{rgb}{0.21,0.49,0.74}
\usepackage[breaklinks,colorlinks,allcolors=cvprblue]{hyperref}

\def\paperID{*****} 
\def\confName{CVPR}
\def\confYear{2026}

\title{NightSteps: A Night-Time Dataset and Benchmark for \\Pedestrian Crossing Intention Prediction}

\author{Sjoerd Takken\\
University of Amsterdam\\
\and
Shaodi You\\
University of Amsterdam\\
}

\begin{document}
\maketitle
\begin{abstract}
Autonomous vehicles use pedestrian crossing intention prediction (PCIP) to participate safely in traffic. Existing models are largely trained under favourable daytime conditions with sufficient daylight. Little is known about the robustness of these models under night-time and low-light conditions, due to the lack of datasets containing night-time footage. This paper investigates how a dataset, named NightSteps, can be constructed and evaluated as a baseline benchmark. The dataset is based on nighttime dashcam footage collected from six different cities around the world, capturing a wide variety of road conditions. The dataset consists of dashcam-recorded point-of-view footage, raw data of pedestrians’ locations with corresponding ‘track\_ids’, and ground-truth labels containing accurate information about pedestrian crossing behaviour, resulting in a diverse dataset. Gemini, ChatGPT, and human assessment are used to demonstrate the usability of the dataset by predicting crossings in clips from the dataset and comparing the results with the initial ground-truth labels. These results show that human input achieves an accuracy of 88.17\% and an F1-score of 80.21\%. Predictions from Gemini and ChatGPT perform lower, with accuracies of 60\% and 79.13\% respectively, and F1-scores of 42.50\% and 63.64\%. The evaluation demonstrates that the NightSteps dataset can be used for future model evaluation to test the performance of PCIP, with the aim of improving road safety at night. The complete source code is publicly available on GitHub.\footnote{\url{https://github.com/sjoerdjt/NightSteps_dataset}}

\end{abstract}

\section{Introduction}
\label{sec:intro}

Pedestrians are recognized as the most vulnerable group in road traffic, as they lack physical protection \parencite{neamțu2024safety}. The unpredictable behaviour and sudden movements of pedestrians can have serious consequences for the surrounding traffic, as well as for the pedestrians themselves. This is concerning as research has shown that drivers' reaction times are often insufficient for effective braking before a potential collision becomes unavoidable  \parencite{drozdziel2020drivers}. To address this issue, an increasing number of mandatory safety features such as sensors or automatic braking systems can contribute to road safety, extending all the way to fully autonomous cars. In the Netherlands, it has recently become legal to allow a car to drive itself completely \parencite{nos-nieuws-2026}. This is supported by research from the Dutch Vehicle Authority (RDW), which carried out an eighteen-month study on fully autonomous vehicles \parencite{RDW}. The findings indicates that fully autonomous vehicles have a positive contribution to road safety.

Autonomous vehicles use artificial intelligence (AI) systems to detect road users. One of the most commonly used detection methods is the You Only Look Once (YOLO) computer vision model from Ultralytics. This model uses a single convolutional neural network (CNN) that predicts bounding boxes and assigns them to a specific class \parencite{ultralytics-2026}. Pedestrians can also be effectively detected and tracked by this model to monitor the pedestrians physical behaviour. However, only detecting a pedestrian is not enough to anticipate the pedestrians behaviour. For safe decision-making, a prediction is needed to determine what a pedestrian is likely to do next, also known as pedestrian intention prediction \parencite[PIP;][]{azarmi2025pedestrian}. In this thesis, the focus is specifically on crossing behaviour, known as pedestrian crossing intention prediction (PCIP). This prediction is important to determine whether someone is going to cross the street, wait, speed up or slow down. An autonomous vehicle must be able to assess in advance whether someone is going to cross, making PCIP essential.

Several studies have focused on PCIP. A study by \textcite{azarmi2025pip_2} makes predictions up to four seconds in advance in busy traffic situations using the PIP-Net model, which takes into account both kinematic data and spatial features. Another study by \textcite{azarmi2025pedestrian} expands on this prediction by integrating multimodal data through prompt templates. Here, contextual information consisting of visual frames and vehicle dynamics is integrated into prompts, after which the Vision-Language Foundation Model (VLFM) can make an even more precise prediction. Furthermore, study by \textcite{xu2024pedestrian_main} demonstrates that this intention prediction is also highly dependent on the interaction between pedestrians and cars for its accuracy. This distinguishes the research from others that do not incorporate this interaction into the prediction.

Although research on PCIP has advanced significantly, a problem persists regarding how the models are trained. The mentioned studies demonstrate high prediction performance but have been trained on daytime data, where detection and prediction are easier. It is still uncertain whether models trained on daytime data will also function effectively in night-time conditions with the absence of daylight.

To improve road safety for both autonomous vehicles and pedestrians, it is essential that AI models are able to make accurate predictions. Due to the limited availability of night-time datasets, little is currently known about how existing models can predict pedestrian crossing intention in low-light environments with shadows, motion blur, and poor visibility of body posture and walking direction.

How reliable are PCIP models when daylight disappears? This research is therefore relevant for minimizing the gap between daytime and night-time by constructing a baseline benchmark dataset for evaluating model performance under low-light conditions. The dataset provides a foundation of night-time data, enabling future research to gain insights into the robustness of PCIP models in low-light environments to ensure safety in realistic environments at night.

\section{Related Work}

\subsection{Existing PCIP Methods}

Within the field of pedestrian intention prediction, much research has been carried out on pedestrian crossing behaviour. Existing studies employ various methods that primarily focus on this behaviour, rather than other road users. AI models based on machine learning are intergraded into these methods to make predictions focused on visual and temporal information.

These predictions are presented in research by \textcite{ni2026pcip}, in which the Pedestrian Crossing Intention Prediction (PCIP) model uses multiple perspectives for prediction. In this model, a quasi-polar coordinate system is established between the pedestrian and the vehicle with their relative distance to analyse the interaction. A representation of the ego-vehicle, from whose perspective the footage is recorded, is also constructed to determine the speed relative to the surrounding traffic. Then, a representation of the pedestrian is created to record their actions, poses, and bounding boxes. This allows their behaviour to be observed. Feature encoding processes these features to assign weights using feature fusion. These weights are classified by a Multi-Layer Perceptron (MLP)-based classification into an output that predicts whether the pedestrian will cross. Under optimal conditions with sufficient light and good weather, the model achieves accuracy scores of 89\% and 93\% on the Joint Attention in Autonomous Driving \parencite[JAAD;][]{kotseruba2016joint} and Pedestrian Intention Estimation \parencite[PIE;][]{rasouli2019pie} datasets, respectively.

Significant results have also been achieved in a study from \textcite{xu2024pedestrian_main}. This research introduces the Pedestrian Vehicle Information Modulation (PVIM) model, which achieves an accuracy of 91\%. This is a 2\% improvement over the previous best models Graph+ \parencite{cadena2022pedestrian} and 'Feature Fusion and Spatio-temporal Attention' \parencite[FSTA;][]{yang2022predicting}. This model establishes a link between pedestrians and vehicles based on the pedestrian-vehicle spatial context (PVSC), after which a temporal bilinear attention module (TBAM) is applied to remove temporal redundancy for more accurate predictions. 

Although these models achieve high accuracy in predicting pedestrian crossing behaviour, a major limitation becomes clear. All models are trained on data where both high-quality visual and temporal information is available, specifically in high-light conditions. As a result, a detection model such as YOLOv8 can detect specific body features with high accuracy. This valuable information can be used to make accurate predictions. When visibility decreases due to a lack of light, the models’ performance for prediction may decline significantly. The study by \textcite{azarmi2025pip_2} notes that there are already more misclassifications and temporal inconsistency at night, where pedestrian detection performance decreases and pedestrians may be classified as part of the background. This could lead to incorrect PCIP.

The limitation of the lack of light creates a research gap. Existing models demonstrate the potential to achieve high scores, but it is not yet clear to what extent these models perform well in night-time traffic conditions. The robustness of the models provides insight into the safety of the systems in night-time settings under adverse light conditions.

\subsection{Existing PCIP Datasets}

Many datasets for PCIP are publicly available for testing and training models. The most common datasets are PIE \parencite{rasouli2019pie}, JAAD \parencite{kotseruba2016joint}, Trajectory Inference using Targeted Action priors Network \parencite[TITAN;][]{malla2020titan}, and Frontal Urban-PIP \parencite[FU-PIP;][]{azarmi2025pip_2}. These datasets contain videos with footage from inside the vehicle, other than Closed-Circuit Television (CCTV) data, where footage is collected using wide-angle surveillance cameras, particularly at intersections. 

In addition to the videos, these datasets contain other relevant information needed to make the prediction. The datasets mentioned above include numerical data on the coordinates of bounding boxes. These are boxes placed around a pedestrian using detection models such as YOLOv8 to determine the exact location of a person within a frame. Additionally, the datasets share the feature that IDs are linked to the boxes which makes tracking of a pedestrian across multiple frames possible.

These datasets stand out from other datasets because they include additional information about behavioural intention. They incorporate 'crossing intention confidence' and include text labels for pedestrians’ actions. These labels include "walking," "standing," "looking," "not looking," "crossing," and "not crossing." As a result, the dataset is immediately usable for testing and training a model. With the PIE and FU-PIP datasets, the emphasis is on predicting pedestrians who intend to cross because of the explicit information about crossing intention. The JAAD dataset is more intended for general prediction based on interaction between pedestrian and vehicle, while the TITAN dataset focuses more on long-term prediction based on previous actions.

The data mostly consist of video footage recorded during daytime. The most commonly used dataset for PCIP is PIE, which does not contain a single clip of night-time footage at all. This dataset was recorded entirely during the daytime under sunny or overcast weather conditions \parencite{rasouli2019pie}. The JAAD dataset also consists mainly of daytime footage. Only a few clips were recorded at night (3 out of 347). The TITAN dataset contains ten hours of video footage recorded during the day. Specific quantities of night-time footage are not mentioned. According to the paper by \textcite{azarmi2025pip_2}, the FU-PIP dataset contains the largest amount of night-time data, but to a lesser extent as well. However, this dataset has never been published. 

The lack of low-light data in the datasets creates a research gap. Although the data are suitable for model usage, there are too few night-time images available, which makes it unclear how the models perform on complete night-time datasets.

\subsection{Related Pedestrian Tasks}

Within the field of PCIP, numerous studies have been conducted under general traffic conditions. These studies often highlight limitations that cast doubt on the robustness of the models in challenging situations, which consequently introduces wider complications. Research by \textcite{elgazwy2025predicting} demonstrates that existing models are insufficiently robust against adverse weather conditions, such as rain or fog. Previous models primarily utilise Convolutional Neural Networks (CNNs) or Recurrent Neural Networks (RNNs), which yield poor results under rainy conditions. As a solution, the research by \textcite{elgazwy2025predicting} introduces a transformer model in which poorly visible images obscured by rain are restored. In this process, blurred images are corrected before a pedestrian’s crossing intention is predicted. Results from this research demonstrate that earlier and more accurate detection takes place, enabling earlier predictions to be made. 

Another study by \textcite{ahmed2023multi} shows a limitation and has investigated, within the context of pedestrian crossing prediction, how existing models perform when pedestrians appear small in the frame or are partially obscured by objects such as lampposts or traffic signs. This makes detecting a pedestrian’s posture and movements more difficult, making accurate prediction more challenging. These situations frequently occur in existing datasets such as JAAD and PIE. In this study, the prediction is made using a combination of the pose, tracking and a Long Short-Term Memory (LSTM) model. This is a type of RNN that recognises patterns in sequential data, whereby the model retains important information over long periods \parencite{graves2012long}. Pedestrians located far from the camera are often not detected, meaning existing models disregard them. The combination of the data and the LSTM model achieves an improvement in the accuracy of predicting pedestrians who are located in the distance or are obstructed, compared to other models, for both the JAAD and PIE datasets. The processing time for the entire process is 6.1 milliseconds, making it fast enough for real-time use.

Despite existing research into adverse weather conditions and pedestrians who are difficult to detect, there remains a need for datasets and research in night-time conditions so that models using PCIP can be benchmarked under these conditions and improved in future research.

\section{Method: The NightSteps Dataset}

To systematically evaluate PCIP under low-light conditions, this study introduces a self-generated night-time dataset. The construction of the dataset consists of several consequential steps: data collection, data processing and data labelling. These steps ensure that the dataset can be compiled and is suitable for benchmarking PCIP models under night-time conditions. The constructed dataset in this study is denoted as the \textbf{NightSteps} dataset.

\subsection{Motivation and Problem Definition}

Various related work by \textcite{rasouli2019pie}, \textcite{kotseruba2016joint}, and \textcite{malla2020titan} shows that existing datasets contain little to no usable footage recorded at night. The JAAD dataset contains only three out of 347 clips in night-time, while PIE does not contain a single clip in night-time at all. Due to this shortage of night-time datasets, limited research has been conducted into the performance of PCIP models under night-time conditions. In this paper, night-time and low-light are used interchangeably. These conditions are defined as conditions starting one hour after sunset. This is when the last stage of twilight, the astronomical twilight, occurs with the sun eighteen
degrees below the horizon, meaning that it does not affect the illumination of pedestrians \parencite{fahmi2024zodiacal}. To evaluate the robustness of PCIP models, it is essential to construct datasets under low-light conditions, with the aim of improving road safety at night. Therefore, this section describes how a night-time dataset can be constructed to scale up the number of datasets under low-light conditions. This upscaling is essential to expose the models to a wide variety of situations, thereby preventing overfitting to specific scenarios due to the limited amount of data. The pipeline for constructing the NightSteps dataset is structured as follows:

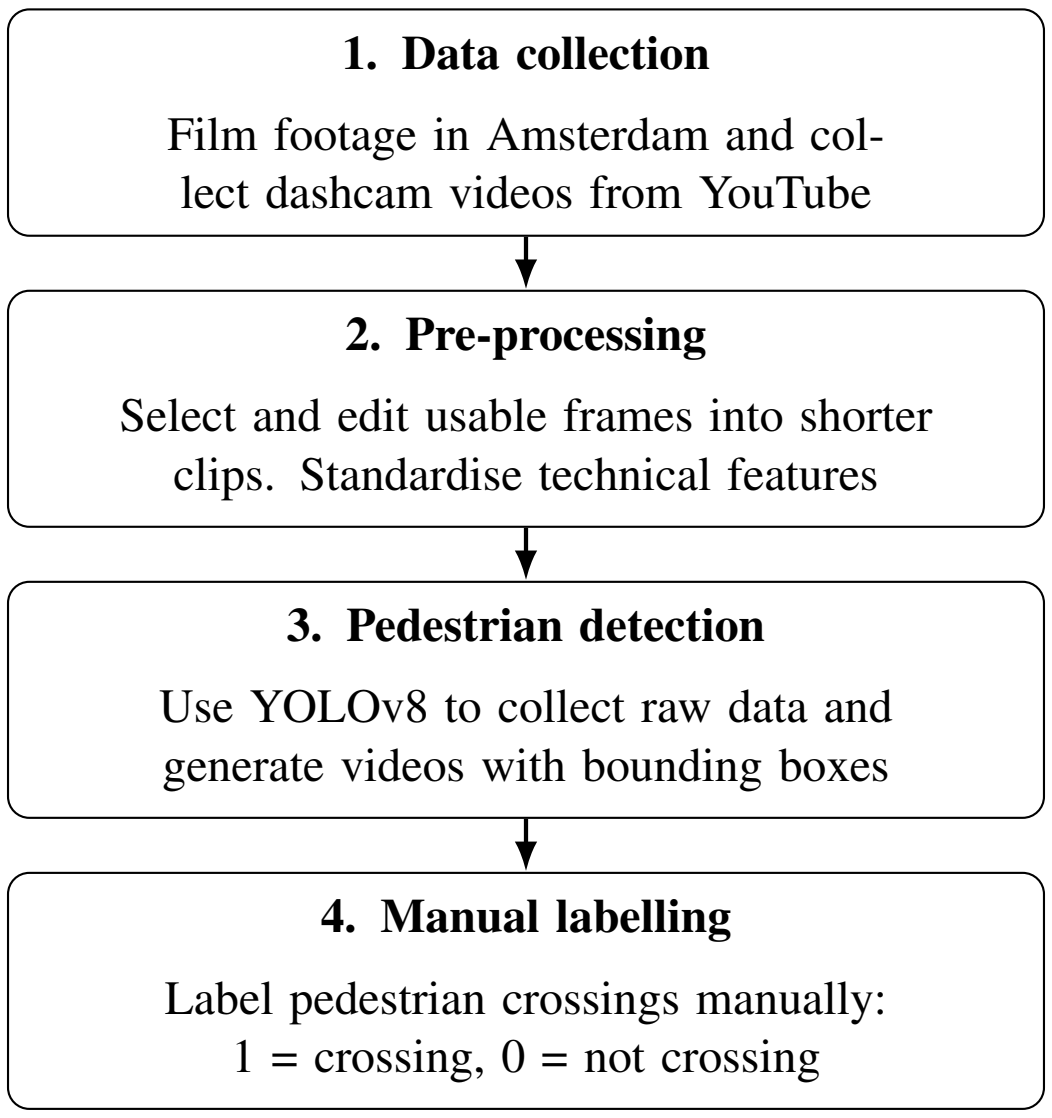
\begin{figure}[H]
    \centering

    \begin{tikzpicture}[
        node distance=0.35cm,
        box/.style={
            draw,
            rounded corners,
            align=center,
            text width=0.78\linewidth,
            inner sep=5pt,
            font=\small
        },
        arrow/.style={
            -{Latex[length=2mm]},
            thick
        }
    ]

    \node[box] (collection) {
        \textbf{1. Data collection}\\[0.15cm]
        Film footage in Amsterdam and collect dashcam videos from YouTube
    };

    \node[box, below=of collection] (preprocessing) {
        \textbf{2. Pre-processing}\\[0.15cm]
        Select and edit usable frames into shorter clips. Standardise technical features
    };

    \node[box, below=of preprocessing] (yolo) {
        \textbf{3. Pedestrian detection}\\[0.15cm]
        Use YOLOv8 to collect raw data and generate videos with bounding boxes
    };

    \node[box, below=of yolo] (labeling) {
        \textbf{4. Manual labelling}\\[0.15cm]
        Label pedestrian crossings manually:\\
        1 = crossing, 0 = not crossing
    };

    \draw[arrow] (collection) -- (preprocessing);
    \draw[arrow] (preprocessing) -- (yolo);
    \draw[arrow] (yolo) -- (labeling);

    \end{tikzpicture}

    \caption{Pipeline for constructing the NightSteps dataset}
    \label{fig:dataset_pipeline}
\end{figure}

\subsection{Data Collection}

\subsubsection{Self-collected Data}

The self-collected part of the dataset consists of footage recorded from a car-mounted dashcam in Amsterdam, the Netherlands. Amsterdam is a busy and dynamic city with a high concentration of cyclists and pedestrians, with a 2024 study showing that 30\% of residents travel on foot and 34\% by bicycle \parencite{verkeer_ams}. The city has many cycle paths, pedestrian crossings and busy junctions where many pedestrians are present, while activity continues throughout the night. The footage of the dataset was recorded at night, with pedestrians being (partially) visible due to artificial lighting.

The visual material consists of a one-hour recording in the centre of Amsterdam. The camera was mounted on the front of the car, creating a point-of-view (POV) image. This method of collecting footage is in line with the way autonomous vehicles gather data and is more effective than using a CCTV camera, which captures only a single junction. This static CCTV viewpoint results in a lack of diversity in traffic situations. As 80\% of the city maintains a speed limit of 30 km/h, pedestrians remain in view for longer, resulting in longer detection times \parencite{klein}.

The footage was recorded between 10:30 pm and 11:30 pm on 14 May 2026. At this time of the day, pedestrians are less visible than in optimal daylight conditions. The images in the dataset are usable and relevant to the research because of their realistic traffic situations and their low-light environment representation, which is in line with the aim of this study.

\begin{figure}[H]
    \centering

    \begin{minipage}[t]{0.315\linewidth}
        \centering
        \includegraphics[width=\linewidth]{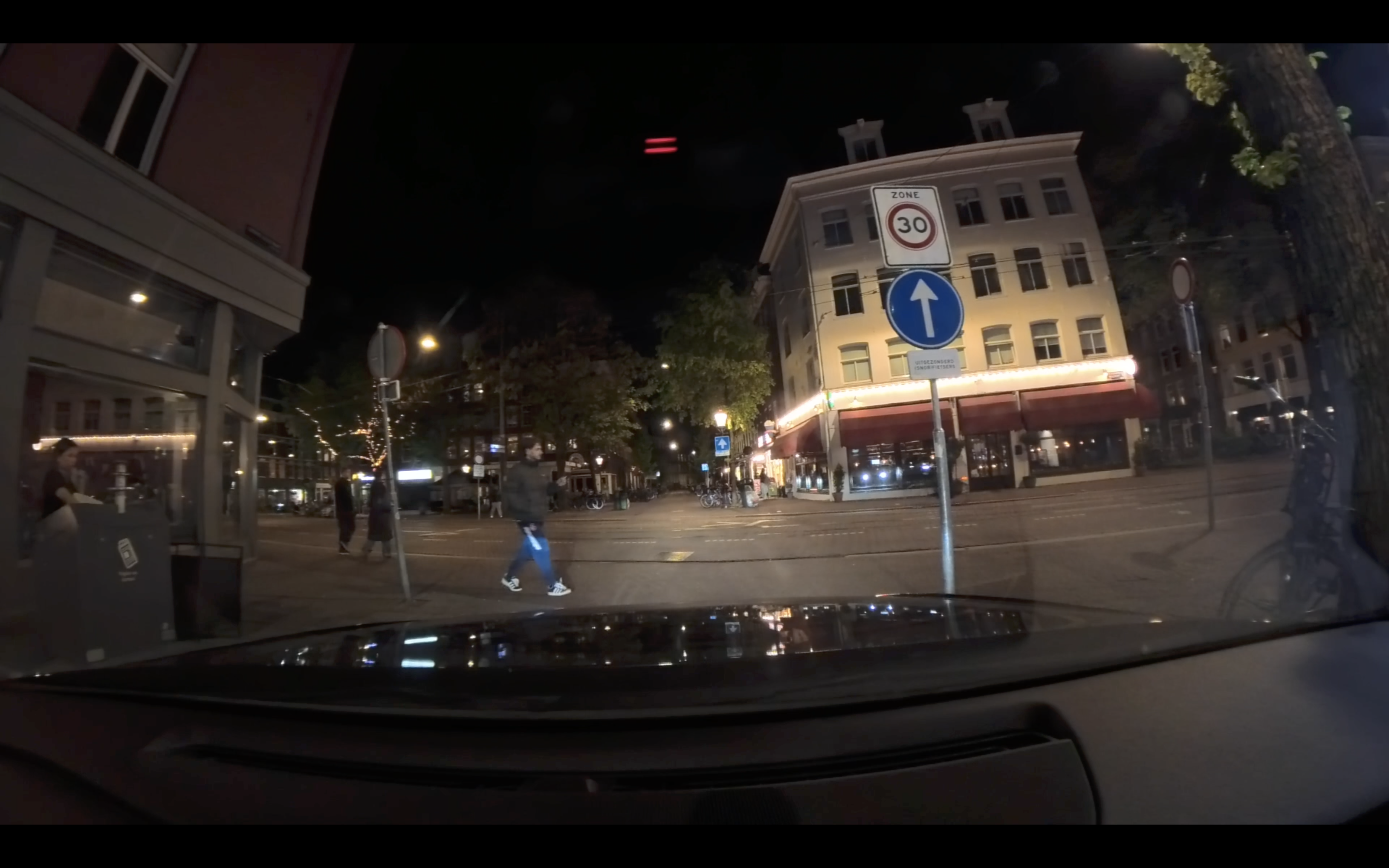}
    \end{minipage}
    \hfill
    \begin{minipage}[t]{0.315\linewidth}
        \centering
        \includegraphics[width=\linewidth]{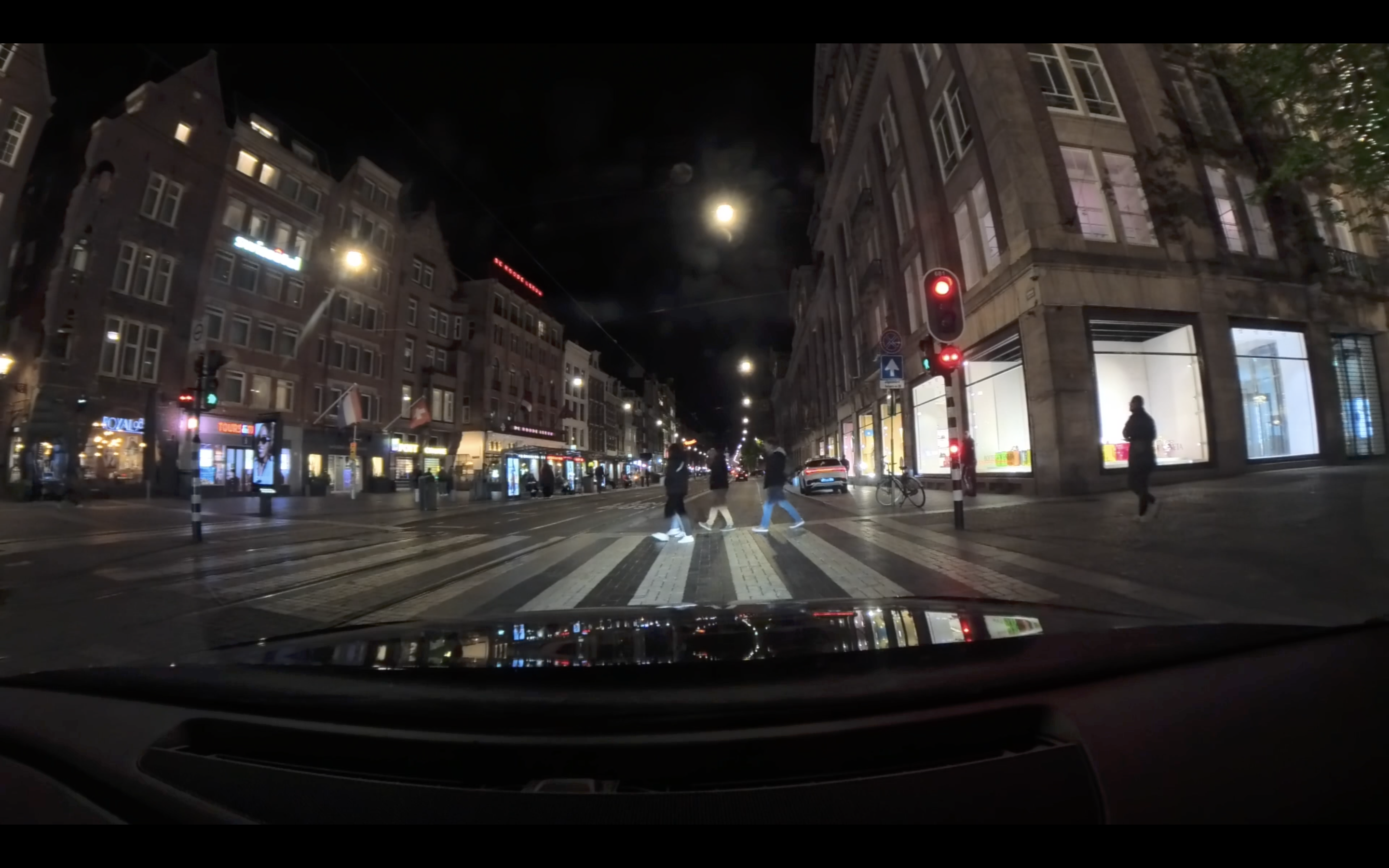}
    \end{minipage}
    \hfill
    \begin{minipage}[t]{0.315\linewidth}
        \centering
        \includegraphics[width=\linewidth]{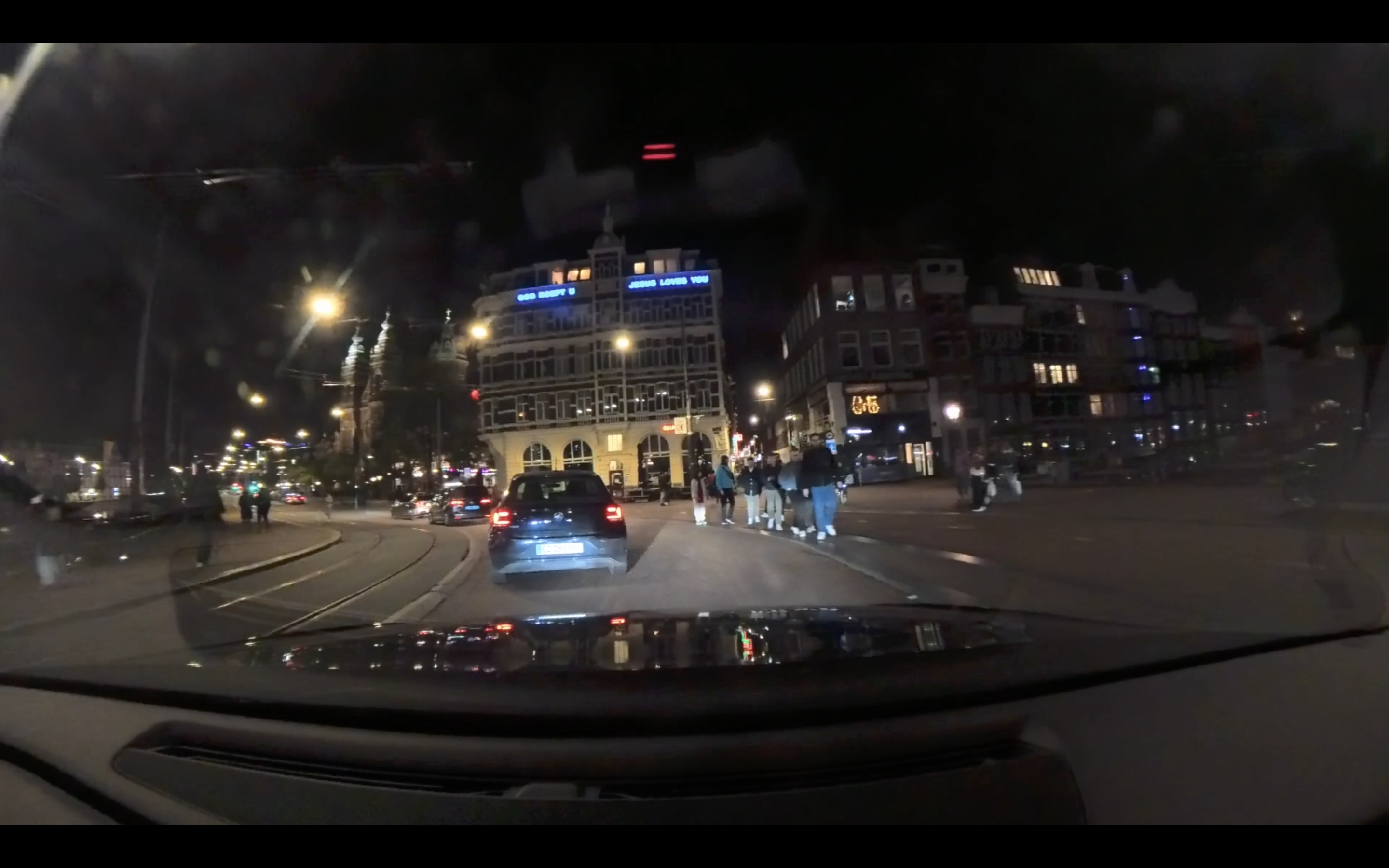}
    \end{minipage}

    \caption{Examples of night-time situations from the self-recorded footage included in the NightSteps dataset}
    \label{fig:amsterdam_examples}
\end{figure}

\subsubsection{YouTube Data}

In addition to the self-collected video material, additional footage was collected to increase the number of usable pedestrians. To this end, additional YouTube videos from five different cities in night were used: New York \parencite{NYC-2023}, Buenos Aires \parencite{BA-2020}, Hong Kong \parencite{Hong_Kong-2020}, Tokyo \parencite{Tokyo-2023} and Madrid \parencite{Madrid-2021}.  The videos feature more pedestrians than the self-recorded video and contain urban traffic situations similar to those in Amsterdam, but on a much larger scale. 

These videos are relevant to the research because they highlight the number of pedestrians and show a more diverse picture of pedestrians across different cities. Not only can pedestrian behaviour differ, but the lighting and the layout of crossings can also vary significantly. This approach results in a more broadly composed dataset that is less dependent on a specific location.

\begin{figure}[H]
    \centering

    \begin{minipage}[t]{0.315\linewidth}
        \centering
        \includegraphics[width=\linewidth]{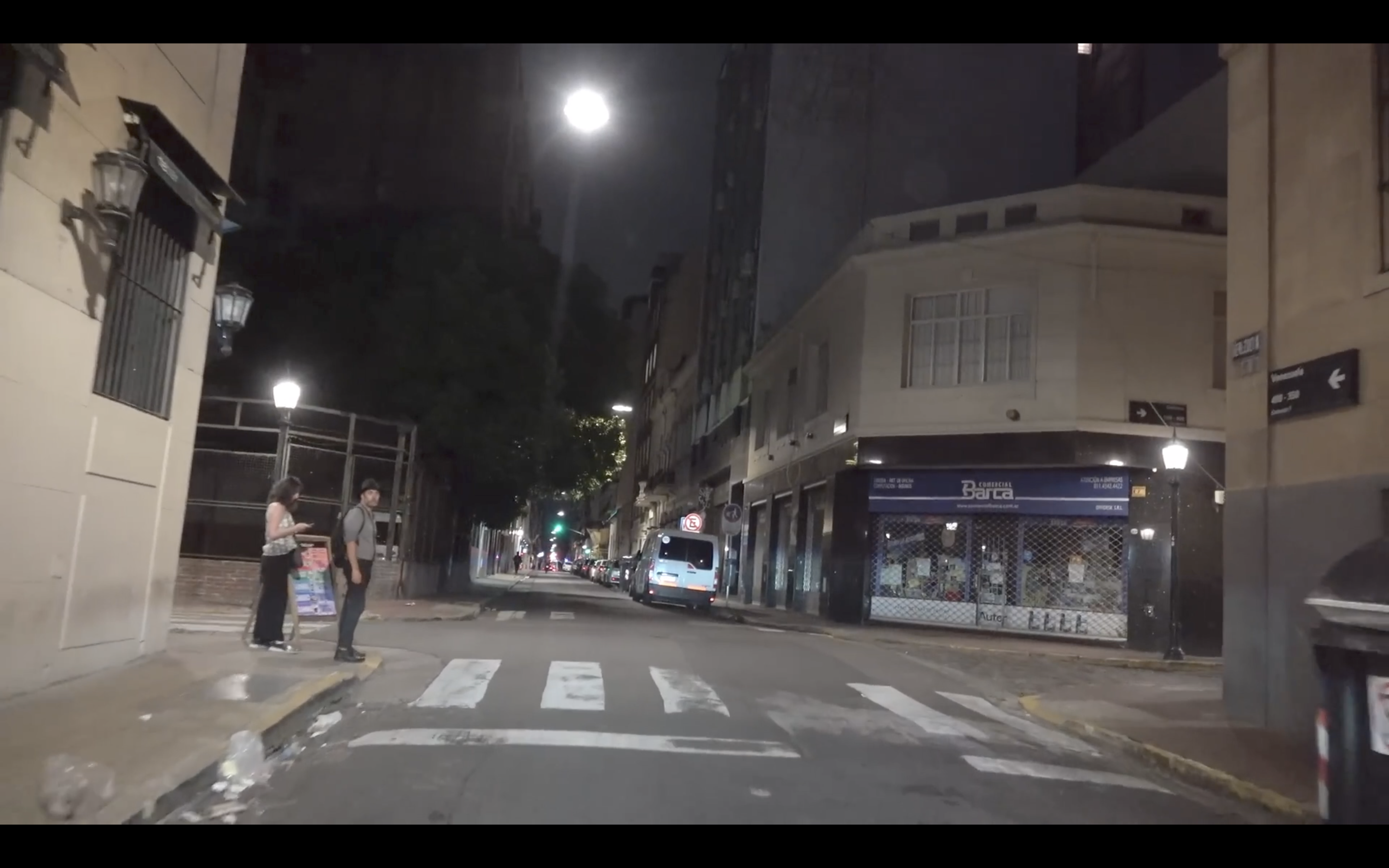}
    \end{minipage}
    \hfill
    \begin{minipage}[t]{0.315\linewidth}
        \centering
        \includegraphics[width=\linewidth]{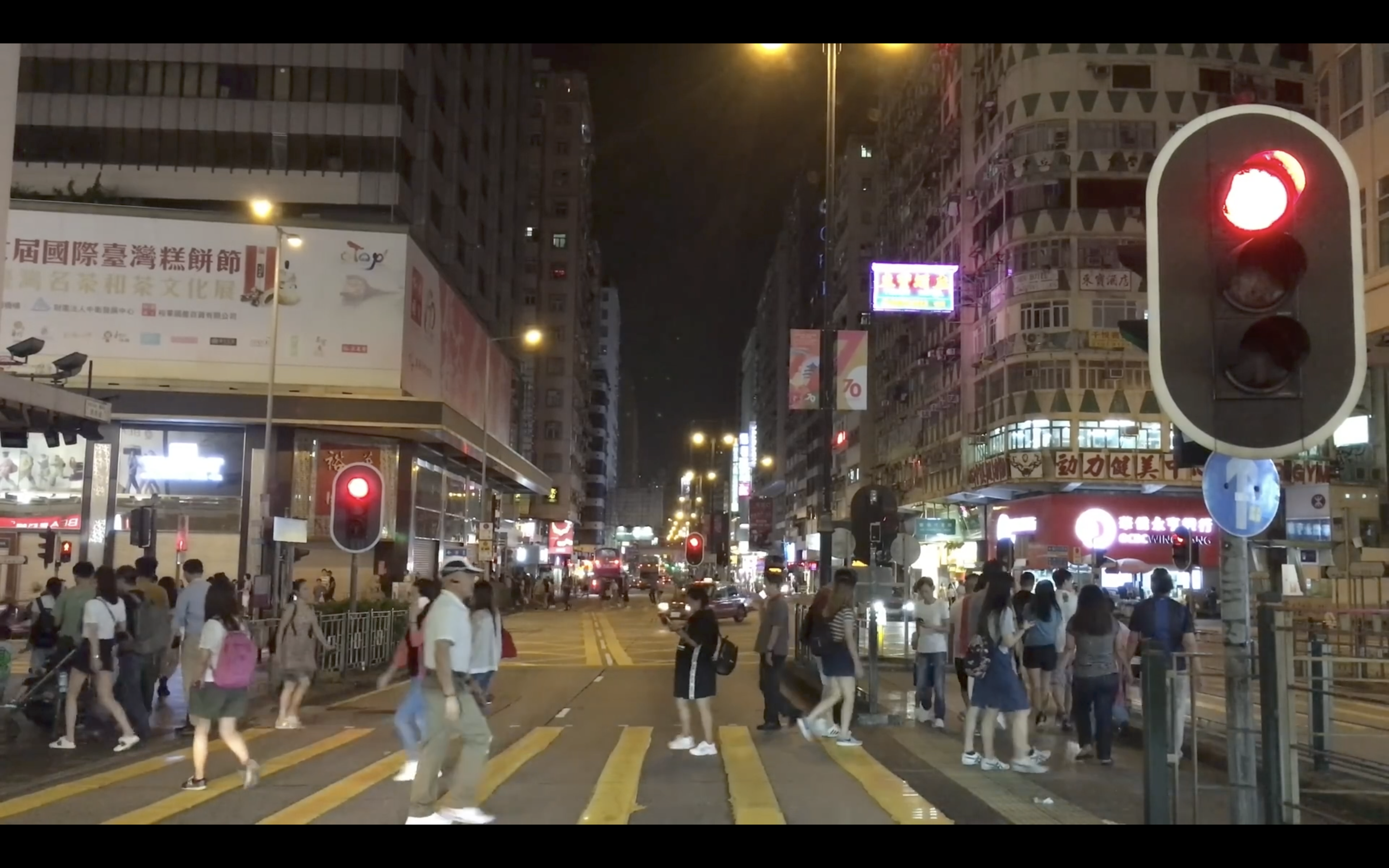}
    \end{minipage}
    \hfill
    \begin{minipage}[t]{0.315\linewidth}
        \centering
        \includegraphics[width=\linewidth]{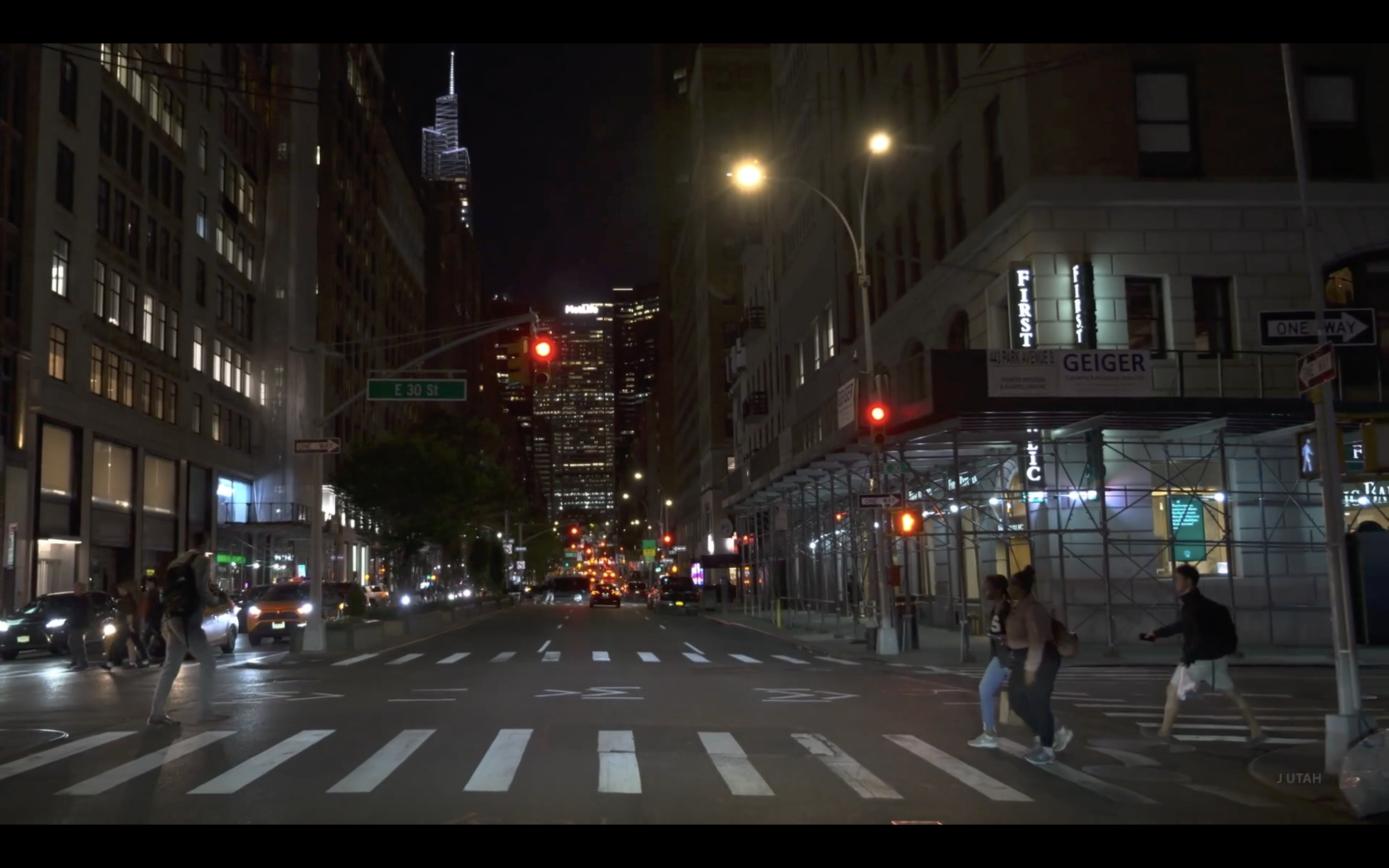}
    \end{minipage}

    \caption{Examples of night-time situations from Buenos Aires (left), Hong Kong (middle), and New York City (right) included in the NightSteps dataset}
    \label{fig:youtube_city_examples}
\end{figure}

\subsection{Technical Features}

The footage in the self-recorded part of the dataset was captured using an Insta360 X5 camera \parencite{Insta360}. This camera features two lenses to create a 360-degree image. To capture the vehicle’s perspective, only the front-facing lens was used, with a wide 170°  view, enabling a broad view of the street to be recorded. The recordings have a resolution of 1920 x 1080 pixels with a ratio of 16:9 and a frame rate of 30 frames per second. The original video consists of two clips of video data in .mp4 format, combined 01:00:42. This entire video will later be edited down to a clip containing only the usable segments from which useful data can be extracted.

The remaining data has been collected from videos with different durations. There is no information available about the use of the cameras, except that they are mounted in the car as dashcams. The .mp4 format is used here as well. Main features are listed in the table below:

\begin{table}[H]
    \centering
    \caption{Overview of the selected video footage}
    \label{tab:video_footage_original}

    \resizebox{\linewidth}{!}{%
        \begin{tabular}{lccc}
            \toprule
            \textbf{City} & \textbf{Year} & \textbf{Length} & \textbf{Resolution} \\
            \midrule
            Amsterdam      & 2026 & 01:00:42 & 1920 x 1080 \\
            Tokyo          & 2023 & 01:05:07 & 3840 x 2160 \\
            Hong Kong      & 2020 & 00:50:07 & 3840 x 2160 \\
            New York City  & 2023 & 01:32:37 & 3840 x 2160 \\
            Madrid         & 2021 & 00:49:19 & 3840 x 2160 \\
            Buenos Aires   & 2020 & 01:22:01 & 3840 x 2160 \\
            \bottomrule
        \end{tabular}%
    }
\end{table}

\subsection{Pre-processing Data}

For the most part, the full videos contain periods in which no pedestrians are visible. As a result, the videos were edited into shorter clips containing only usable footage in which pedestrians are clearly visible. This is essential for reducing computational load. To clean up the data, the software iMovie was used, which allows individual clips to be spliced together \parencite{AppleiMovie2026}. The original YouTube videos contain 4K resolutions and different frame rates. For this study, the data was converted after editing to a standard format, resulting in videos with a resolution of 1920 x 1080 pixels, a 16:9 aspect ratio and a frame rate of 30 frames per second.

\begin{table}[H]
    \centering
    \caption{Overview of the video footage after editing}
    \label{tab:video_footage_edited}

    \resizebox{\linewidth}{!}{%
        \begin{tabular}{lccc}
            \toprule
            \textbf{City} & \textbf{Length} & \textbf{Resolution} & \textbf{Frame rate} \\
            \midrule
            Amsterdam      & 00:12:00 & 1920 x 1080 & 30 fps \\
            Tokyo          & 00:11:24 & 1920 x 1080 & 30 fps \\
            Hong Kong      & 00:15:16 & 1920 x 1080 & 30 fps \\
            New York City  & 00:13:12 & 1920 x 1080 & 30 fps \\
            Madrid         & 00:05:59 & 1920 x 1080 & 30 fps \\
            Buenos Aires   & 00:12:00 & 1920 x 1080 & 30 fps \\
            \bottomrule
        \end{tabular}%
    }
\end{table}

\subsection{YOLOv8 Model for Detection}

To detect and extract pedestrians from the video, the You Only Look Once (YOLO) detection model is used. This is a real-time model capable of localising and classifying objects \parencite{vijayakumar2024yolo}. The model performs this localisation based on detection from individual images. Therefore, the video is first split into individual frames. Each frame is divided into smaller grids, after which features can be precisely identified using convolutional layers. The model then generates a bounding box around each detected pedestrian based on its prediction. A confidence score indicates the certainty of the prediction’s accuracy. The confidence threshold was set on 0.35 to filter poor predictions. For each frame in which a pedestrian’s position is predicted, the coordinates of the box, the corresponding confidence score and an ID are stored in a .csv file, as illustrated in Figure 4a. The pedestrian ID is necessary for tracking, where frames with the same ID are grouped together to follow the pedestrian’s path.

Many versions of the YOLO model are further developed to enable faster and more accurate detection. For this study, the YOLOv8 model was chosen, as it is able to make parallel predictions because of the C2f module. In this model, bounding boxes and classification are calculated separately, allowing more efficient detection. Newer versions have been excluded due to limited literature and a lower degree of alignment with the methodology of this study.

\begin{figure}[H]
    \centering

    \includegraphics[
        width=0.95\linewidth
    ]{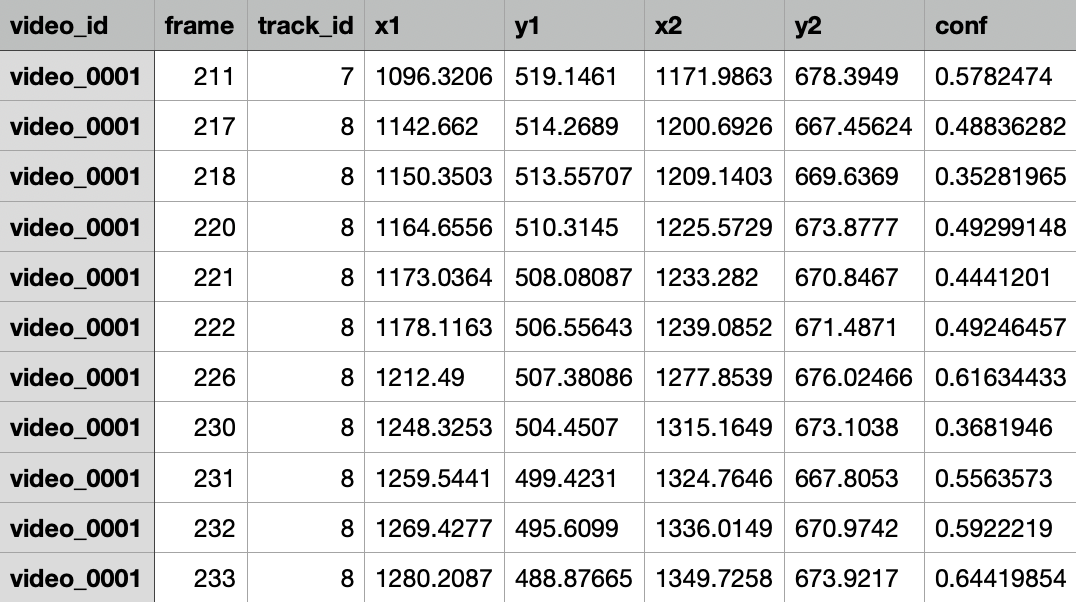}

    \vspace{0.1cm}
    
    {\small (a) Pedestrian detection data stored in a .csv file}

    \vspace{0.3cm}

    \includegraphics[
        width=0.82\linewidth
    ]{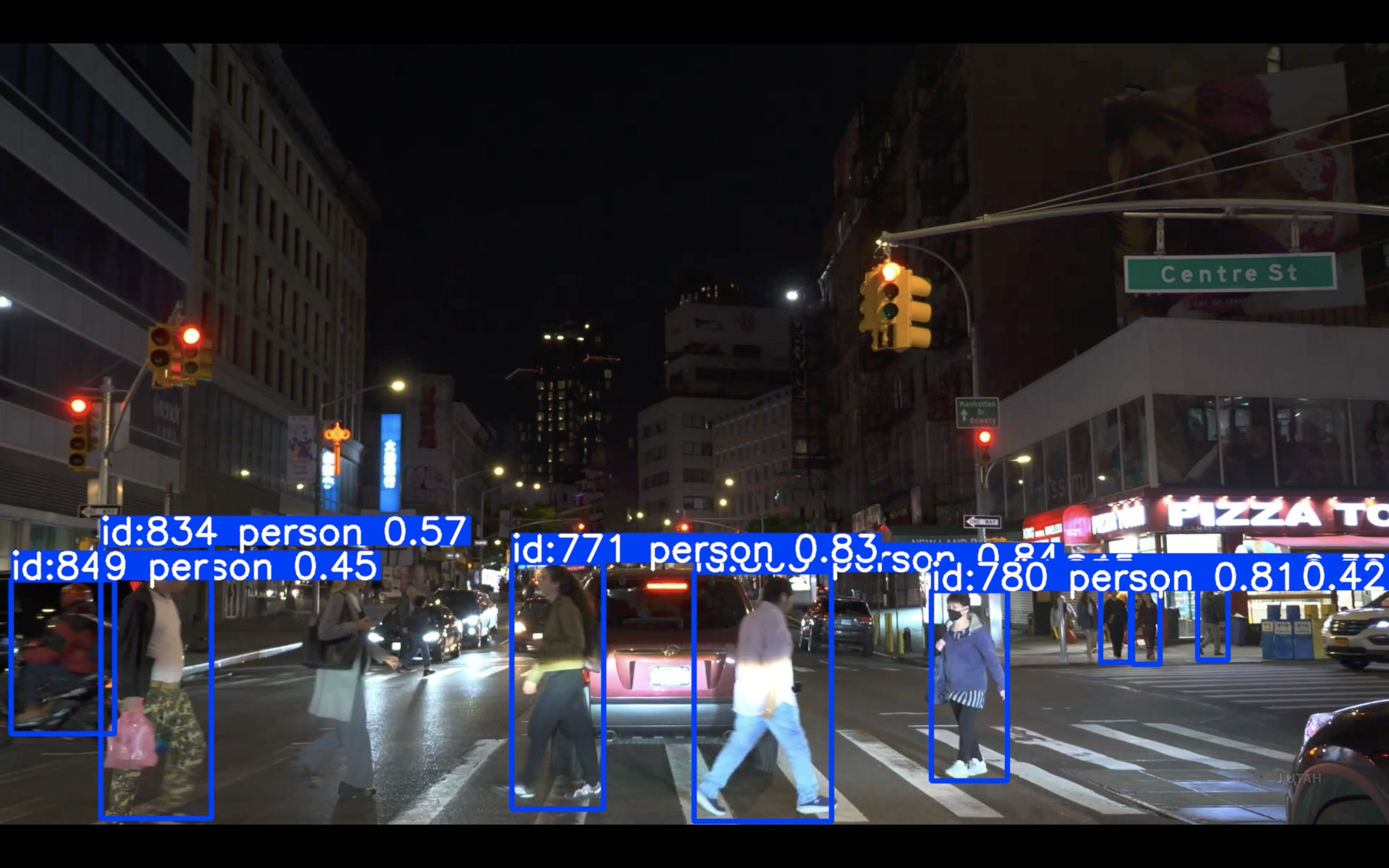}

    \vspace{0.1cm}
    
    {\small (b) Pedestrians detected by YOLO with predicted bounding boxes}

    \caption{Examples of the YOLOv8 pedestrian detection output: (a) detection data stored in a .csv file and (b) visualised pedestrian detections with predicted bounding boxes}
    \label{fig:yolo_detection}
\end{figure}

\subsection{Data Labelling}

Following the detection of pedestrians, usable image frames are used to label the pedestrian’s crossing behaviour. This involves examining behaviour surrounding the crossing process, particularly the seconds immediately before a pedestrian begins to cross. In this study, 'crossing' is defined as the moment at which a pedestrian begins to cross. A pedestrian is labelled as 'not-crossing' when they are not crossing but are clearly visible in the frame. Analysing behaviour before a pedestrian does or does not cross is essential for making a prediction at a later stage.

This crossing is labelled manually by going through the generated frames one by one, noting the corresponding frame number as 'event\_frame' when crossing. Using a binary system, pedestrians in the frames are labelled, with 'crossing' labelled as '1' and 'not crossing' as '0'. Simple versions of prediction models use only this information for the crossing. Advanced models also require the frame numbers marking the end of the crossing when pedestrians cross the road. For a more complete dataset, this extra information has also been included in this study to provide more accurate crossing data. Pedestrians walking parallel to the car on pavements are excluded from the labelling process due to the irrelevance of the pedestrian crossing data. The table below shows a sample of the labelled data, in which 'track\_id', 'crossing', 'start\_frame', 'end\_frame' and 'event\_frame' serve as necessary parameters. The most important parameter for predicting crossing intention is the 'event\_frame'. This is the frame in a video at which the actual event begins, where 'start\_frame' $=$ 'event\_frame' in case of crossing. If the pedestrian does not cross, this 'event\_frame' is set to a frame in which the pedestrian is clearly visible. The model will analyse the surrounding frames to make the prediction.

\begin{figure}[H]
    \centering
    \includegraphics[width=0.85\linewidth]{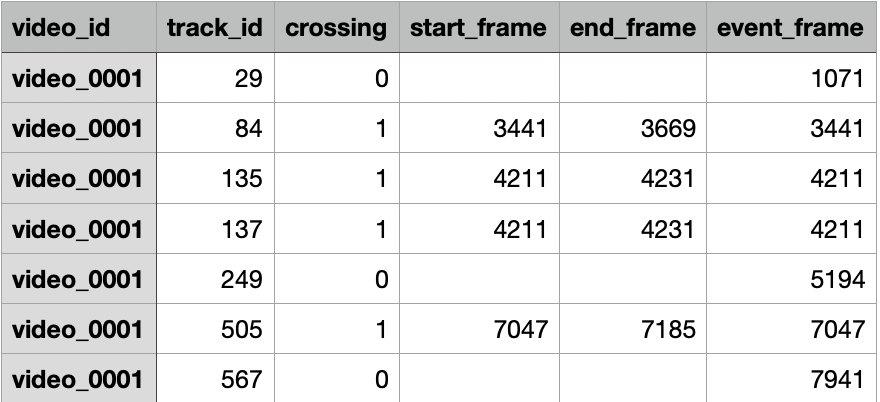}
    \caption{Snippet of crossing labels stored in .csv file}
    \label{fig:crossing_labels}
\end{figure}

\subsection{Data Structure and Format}

The generated dataset consists of a combination of the videos, the corresponding output from the YOLO model, and the assigned labels. This dataset consists of two .csv files per video, with a total of twelve files. The data from the YOLO model provides the coordinates of the bounding boxes, a confidence score, and the corresponding 'track\_id' for the included frames. The data also includes a video showing the bounding boxes around the people, providing a visual representation of the model. The labelling data consists of a 'crossing\_label' with the corresponding 'event\_frame' and information about the 'track\_id' to link the overall detection data to the labels.

For general benchmarking, the original dataset is sufficient because prediction models can use the raw tracking data to predict pedestrian trajectories and crossing intention. To make the dataset accessible for broader evaluation purposes, these pedestrian trajectories are visualised into short, two-second clips leading up to the moment of crossing. These clips contain the bounding boxes predicted by the YOLO model to link the corresponding 'track\_ids' of the pedestrians. This creates, in addition to the raw dataset of detections and labels, a visual representation of the crossing trajectories. These clips allow not only models designed for PCIP to be benchmarked, but also other commonly used AI systems or human assessors. This benchmark is further elaborated in the following chapter.

\begin{table}[H]
    \centering
    \caption{Properties of datasets after YOLOv8 processing}
    \label{tab:dataset_properties}

    \resizebox{\linewidth}{!}{%
        \begin{tabular}{lcc}
            \toprule
             & \textbf{Self constructed} & \textbf{YouTube videos} \\
            \midrule
            Total length of video(s)       & 12:24  & 00:59:33 \\
            No. of frames                   & 22,320 & 107,190 \\
            No. of labelled pedestrians     & 17     & 227 \\
            No. labelled as crossing        & 6      & 55 \\
            No. labelled as not crossing    & 11     & 172 \\
            \bottomrule
        \end{tabular}%
    }
\end{table}

\subsection{Diversity of the Dataset}

Several differences can be observed in diversity between the characteristics of the various cities. First of all, the number of labels in the subsets varies. The subsets for Buenos Aires and Tokyo contain more usable labels than Amsterdam, which has only seventeen. The number of labels in Madrid and Hong Kong is distributed around the average of 40.67 labels. The adjacent table shows the number of usable labels per city.

\begin{table}[H]
    \centering
    \caption{Number of labels per city}
    \label{tab:labels_per_city}

    \begin{tabular}{lc}
        \toprule
        \textbf{City} & \textbf{No. of labels} \\
        \midrule
        Amsterdam      & 17 \\
        Tokyo          & 54 \\
        Hong Kong      & 44 \\
        New York City  & 46 \\
        Madrid         & 34 \\
        Buenos Aires   & 49 \\
        \bottomrule
    \end{tabular}
\end{table}

Furthermore, a difference can be observed in the formation of groups across the different cities. Whereas cities such as Hong Kong, Tokyo and New York City contain many crowded areas where people are relatively close to each other, pedestrians in Madrid, Amsterdam and Buenos Aires are more often seen alone or in groups of two or three. This is partly because the videos showing large groups were recorded in busy locations such as Times Square in New York City. The footage from Amsterdam and Madrid focuses more on smaller streets where fewer people are present. The figure below show differences in group size between the cities.

\begin{figure}[H]
    \centering

    \begin{minipage}[t]{0.315\linewidth}
        \centering
        \includegraphics[width=\linewidth]{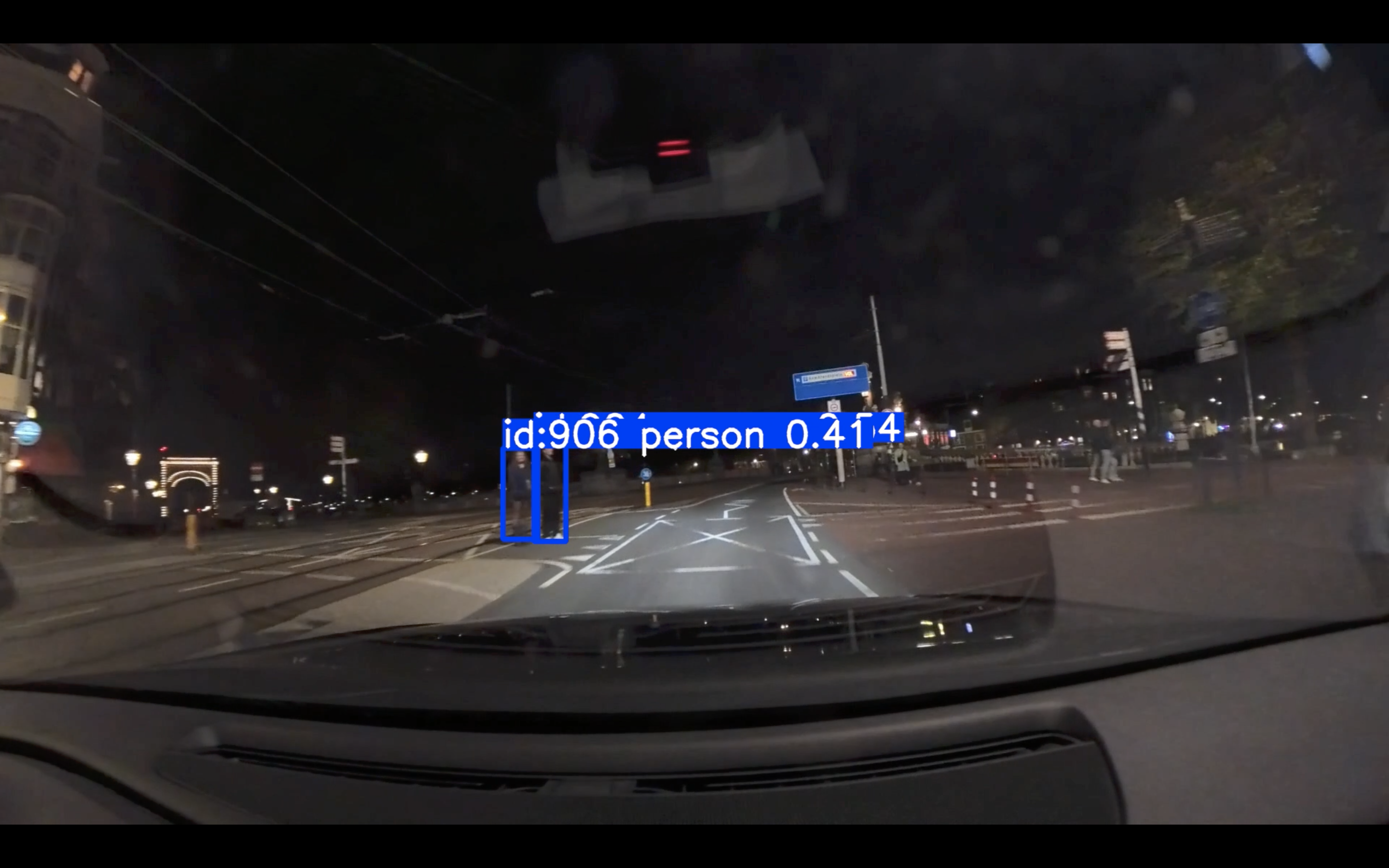}
    \end{minipage}
    \hfill
    \begin{minipage}[t]{0.315\linewidth}
        \centering
        \includegraphics[width=\linewidth]{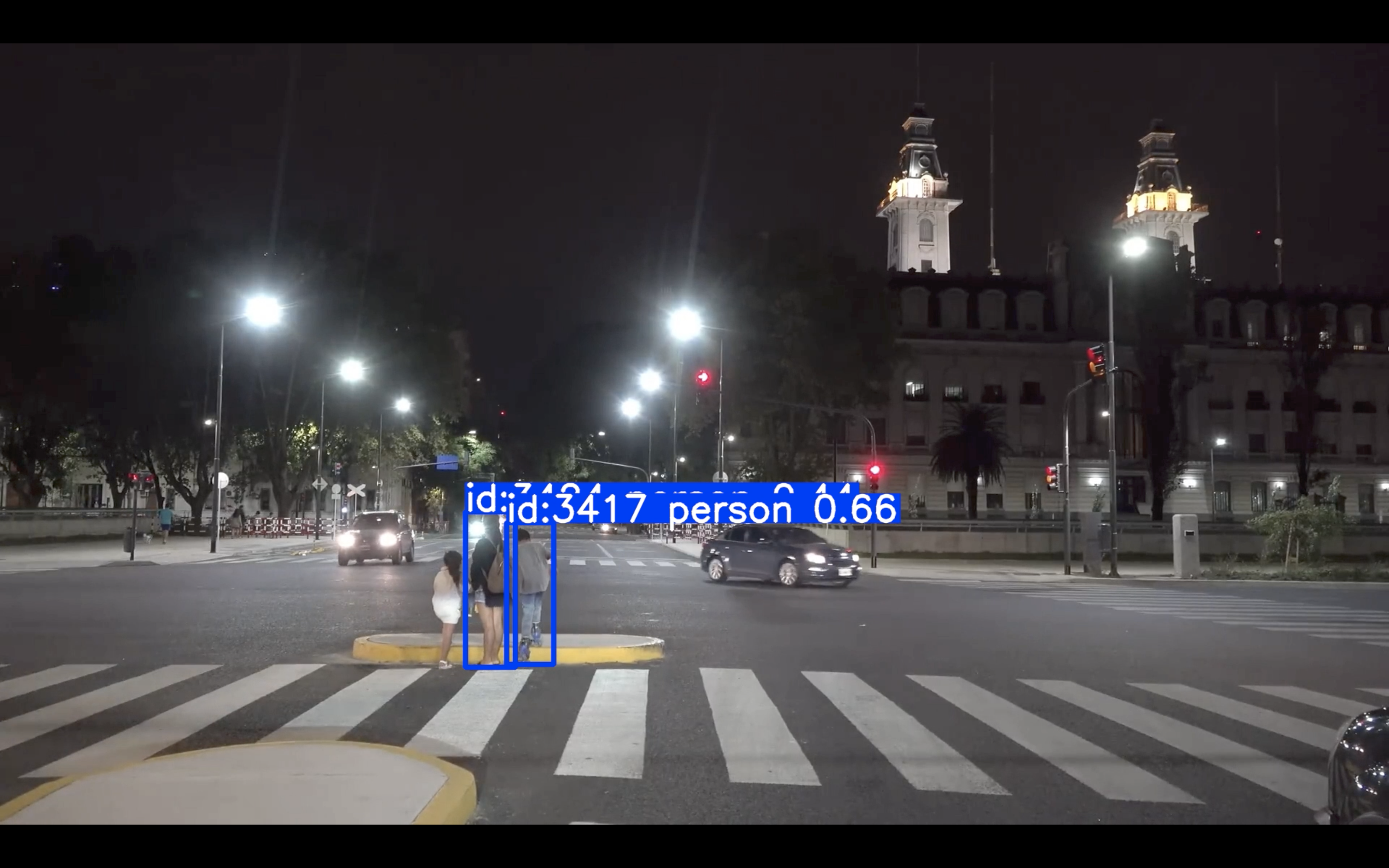}
    \end{minipage}
    \hfill
    \begin{minipage}[t]{0.315\linewidth}
        \centering
        \includegraphics[width=\linewidth]{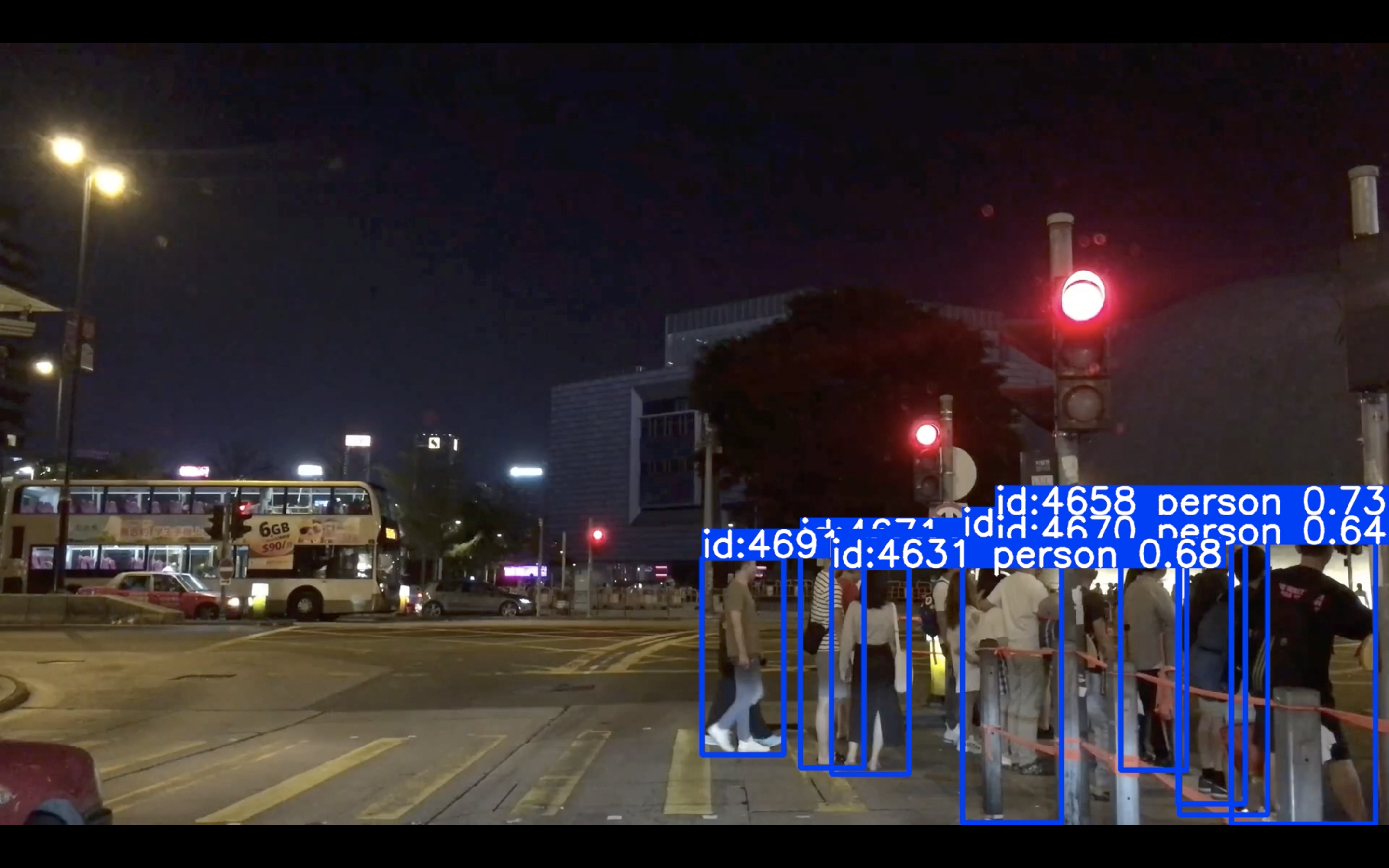}
    \end{minipage}

    \caption{Differences in group size between cities in Amsterdam (left), Madrid (middle), and Hong Kong (right)}
    \label{fig:group_size_examples}
\end{figure}

A difference can also be observed in artificial lighting. In the crowded areas of New York City, Tokyo and Hong Kong mentioned above, there are many billboards, which produce a high level of artificial light. However, this does not mean that pedestrians are detected more clearly here. The overexposure of the screens obscures the rest of the image in certain frames, making the pedestrians harder to detect.

Finally, traffic does not follow the same driving side of the road. In Hong Kong and Tokyo, vehicles drive on the left-hand side of the road, while the other cities drive on the right-hand side. All these differences demonstrate the diversity between the cities, allowing for a wide range of different situations to be captured. This diversity is beneficial for testing the robustness of a model. The figure below shows a number of further examples with images from the various cities, containing two images per city.

The top left image (a) illustrates a situation in which the presence of the billboards creates overexposure, making it more challenging to observe the pedestrians. Image (b) illustrates the same situation with billboards, but in which the boards provide additional lighting for the street. Both clips are from New York City. The next two images (c and d) illustrate situations in Madrid, featuring both busy locations with many pedestrians and quiet streets with few pedestrians. Image (e) illustrates a situation in Buenos Aires where sufficient lighting reveals a large group of pedestrians, and another small well-lit street without any road markings (f). Image (g) illustrates the left-hand side driving in Hong Kong, with another situation (h) in which a pedestrian is not detected due to obstruction by a lamppost. The images from Tokyo (i and j) illustrate crossing area's with high and low pedestrian traffic. Image (j) illustrates a pedestrian using her mobile phone, which may affect the prediction process. Finally, image (k), captured in Amsterdam, illustrates a situation where the pedestrian is not visible due to limited illumination. Image (l) illustrates a situation in which pedestrians are crossing outside a pedestrian crossing. Although the situations may appear similar due to the lack of light, each situation is unique, which highlights the diversity of the clips.

\begin{strip}
\centering


\begin{minipage}[t]{0.205\textwidth}
    \centering
    \includegraphics[width=\linewidth]{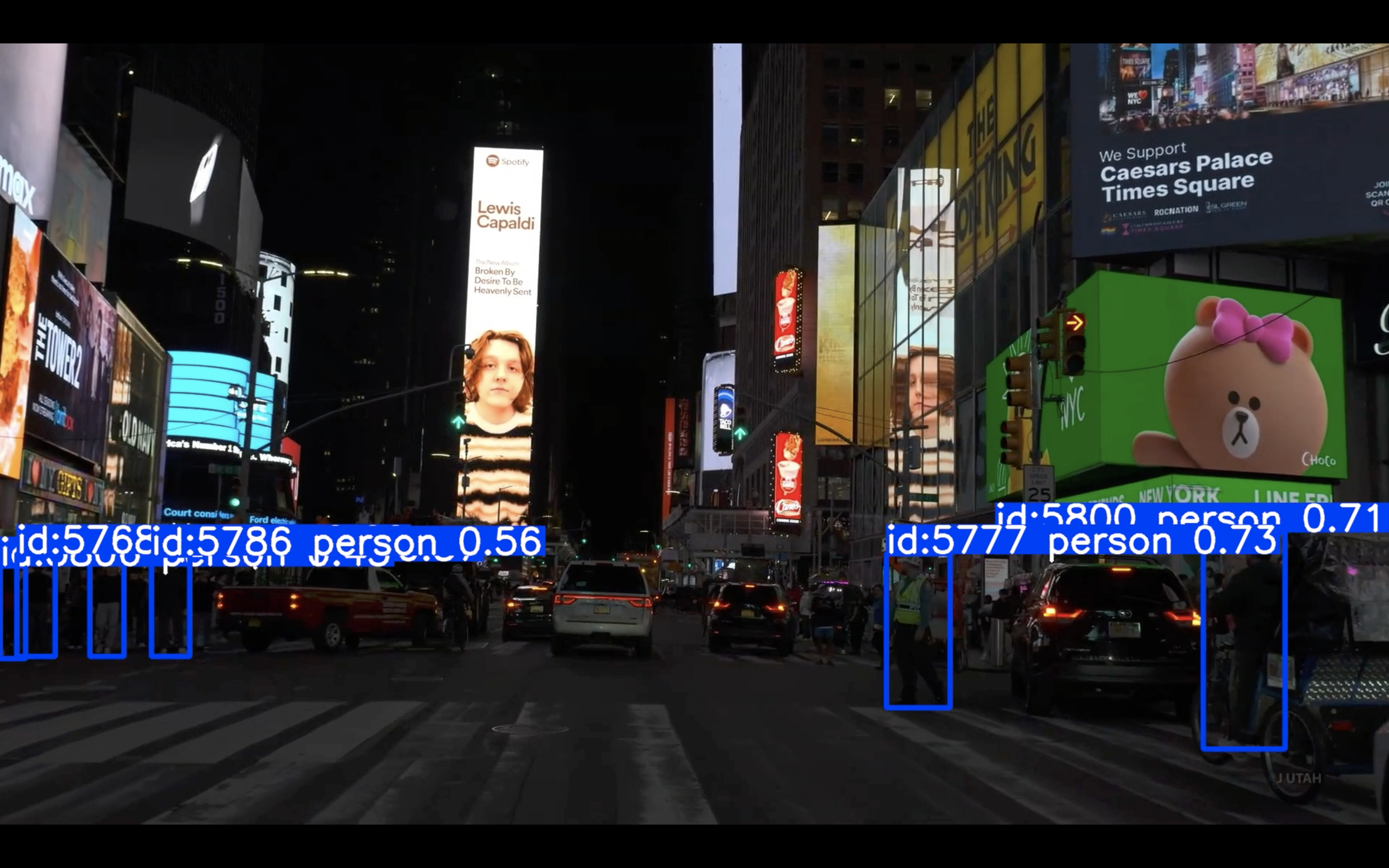}\\[-3pt]
    {\scriptsize (a)}
\end{minipage}
\hfill
\begin{minipage}[t]{0.205\textwidth}
    \centering
    \includegraphics[width=\linewidth]{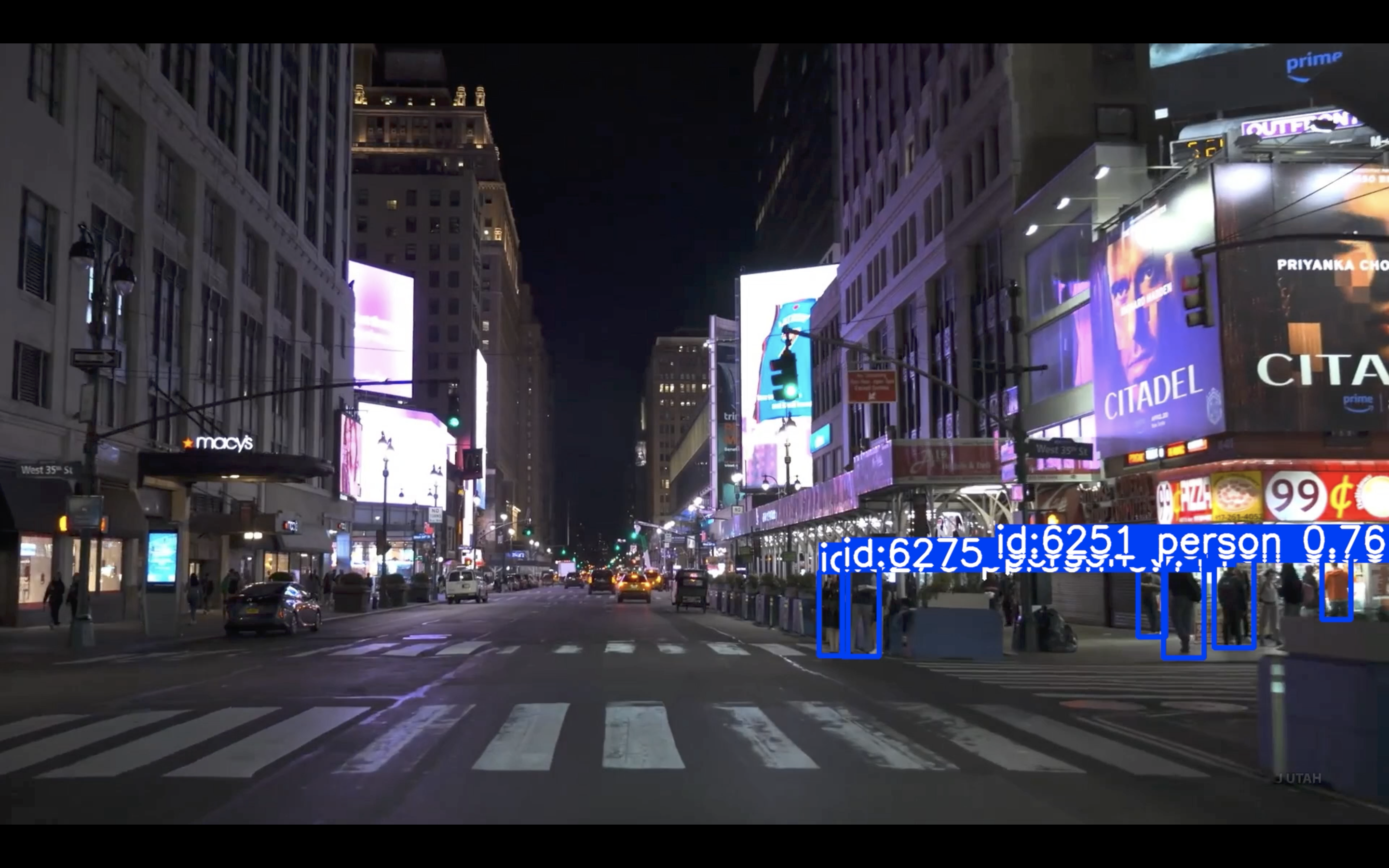}\\[-3pt]
    {\scriptsize (b)}
\end{minipage}
\hfill
\begin{minipage}[t]{0.205\textwidth}
    \centering
    \includegraphics[width=\linewidth]{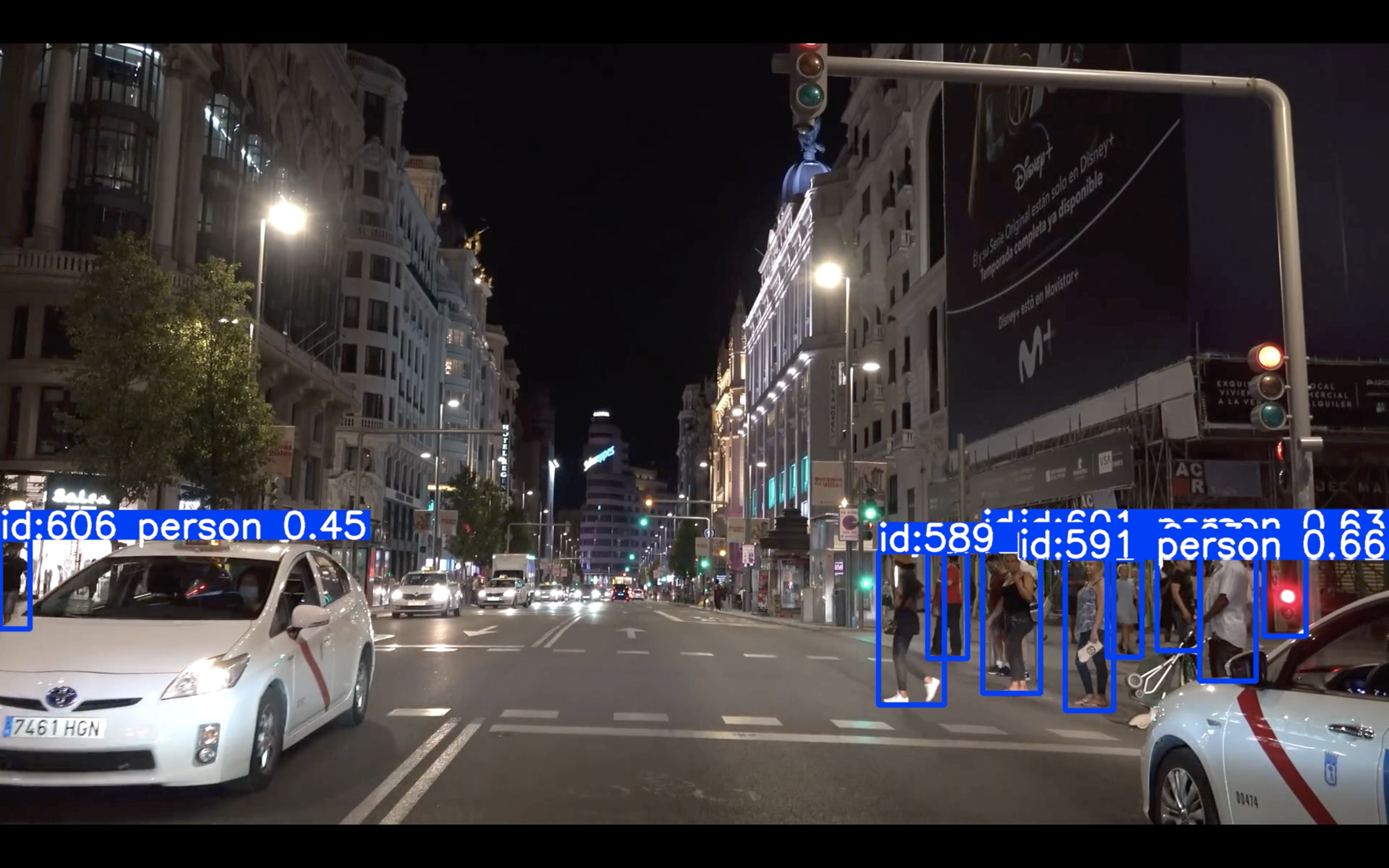}\\[-3pt]
    {\scriptsize (c)}
\end{minipage}
\hfill
\begin{minipage}[t]{0.205\textwidth}
    \centering
    \includegraphics[width=\linewidth]{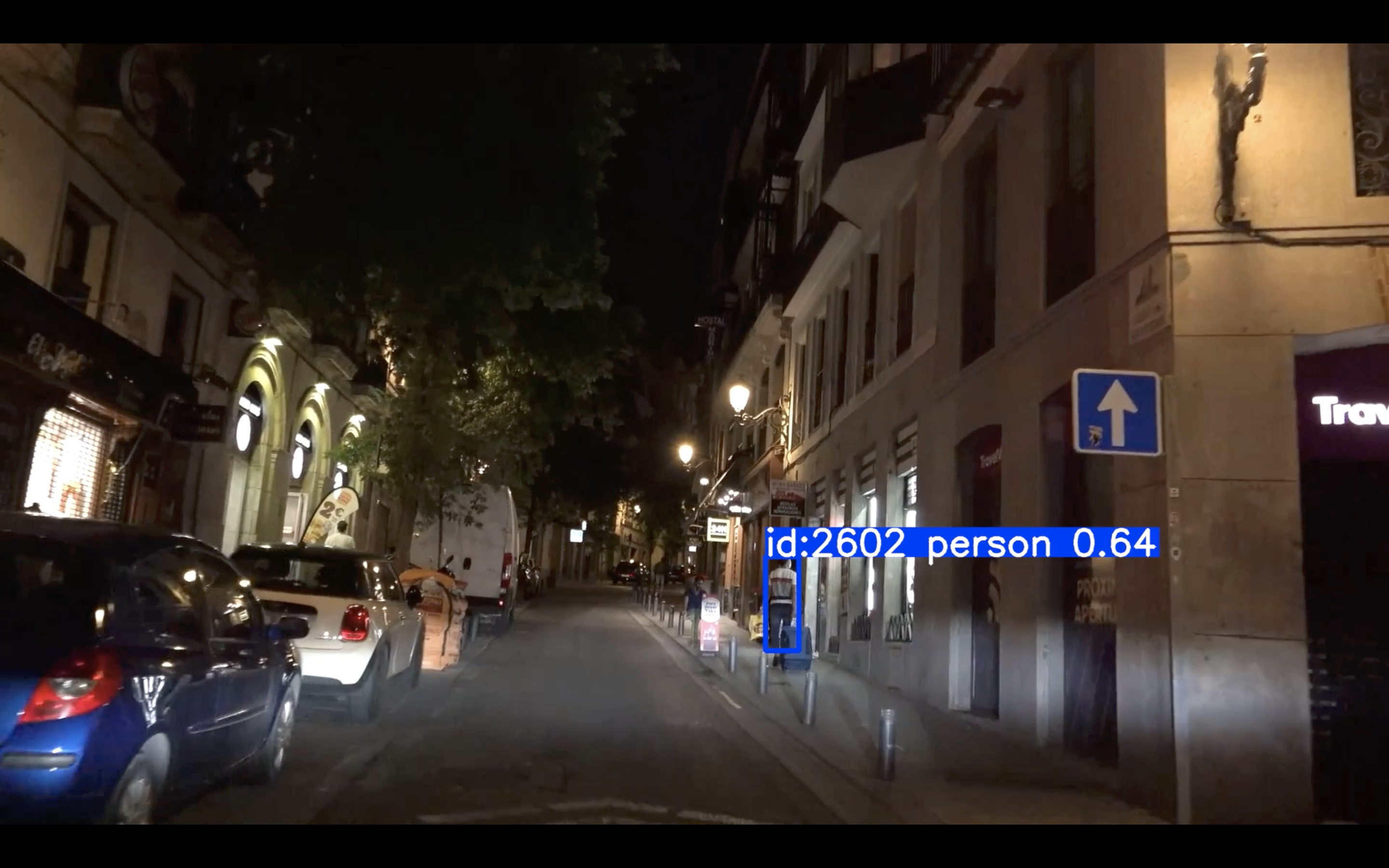}\\[-3pt]
    {\scriptsize (d)}
\end{minipage}

\vspace{2pt}


\begin{minipage}[t]{0.205\textwidth}
    \centering
    \includegraphics[width=\linewidth]{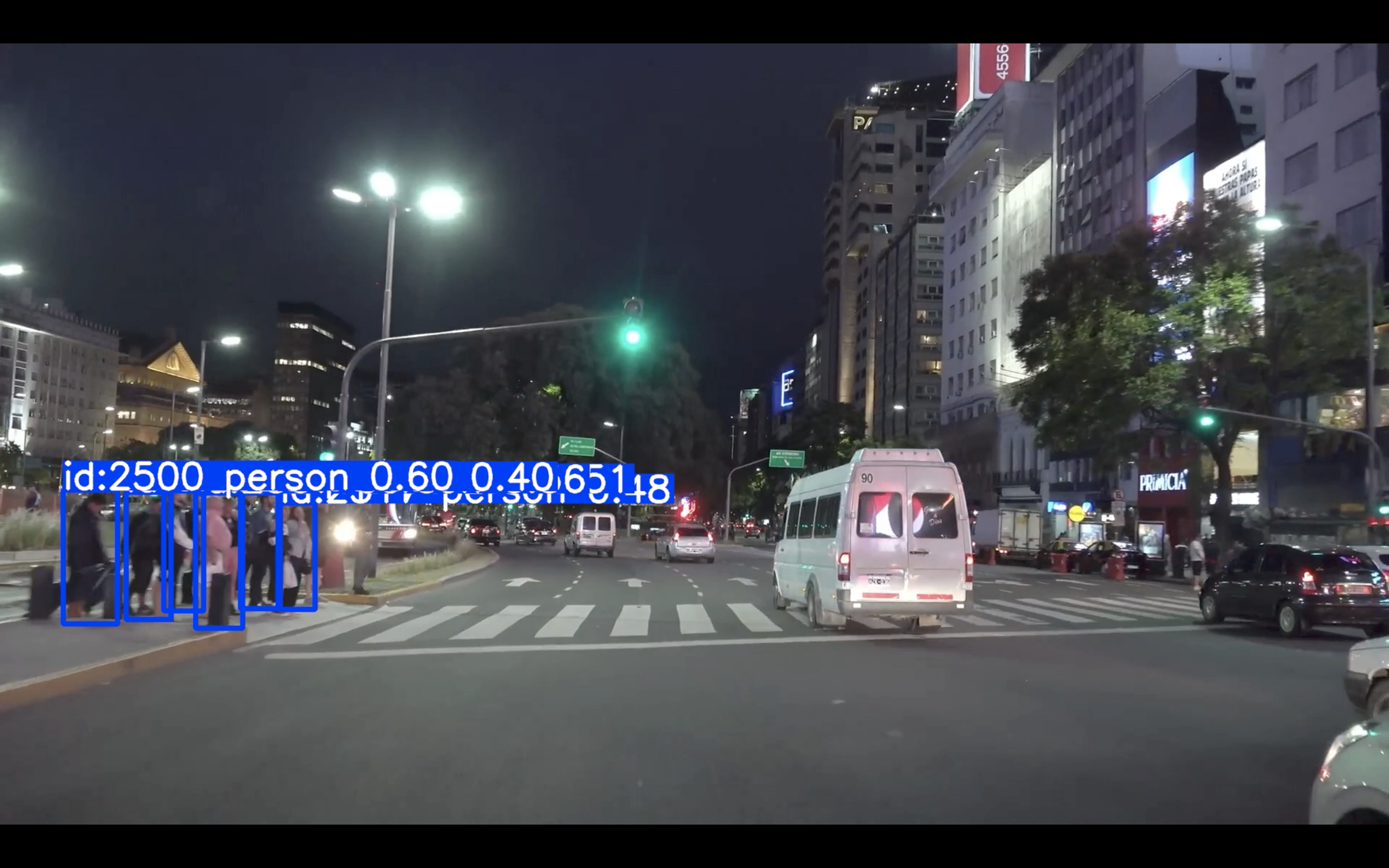}\\[-3pt]
    {\scriptsize (e)}
\end{minipage}
\hfill
\begin{minipage}[t]{0.205\textwidth}
    \centering
    \includegraphics[width=\linewidth]{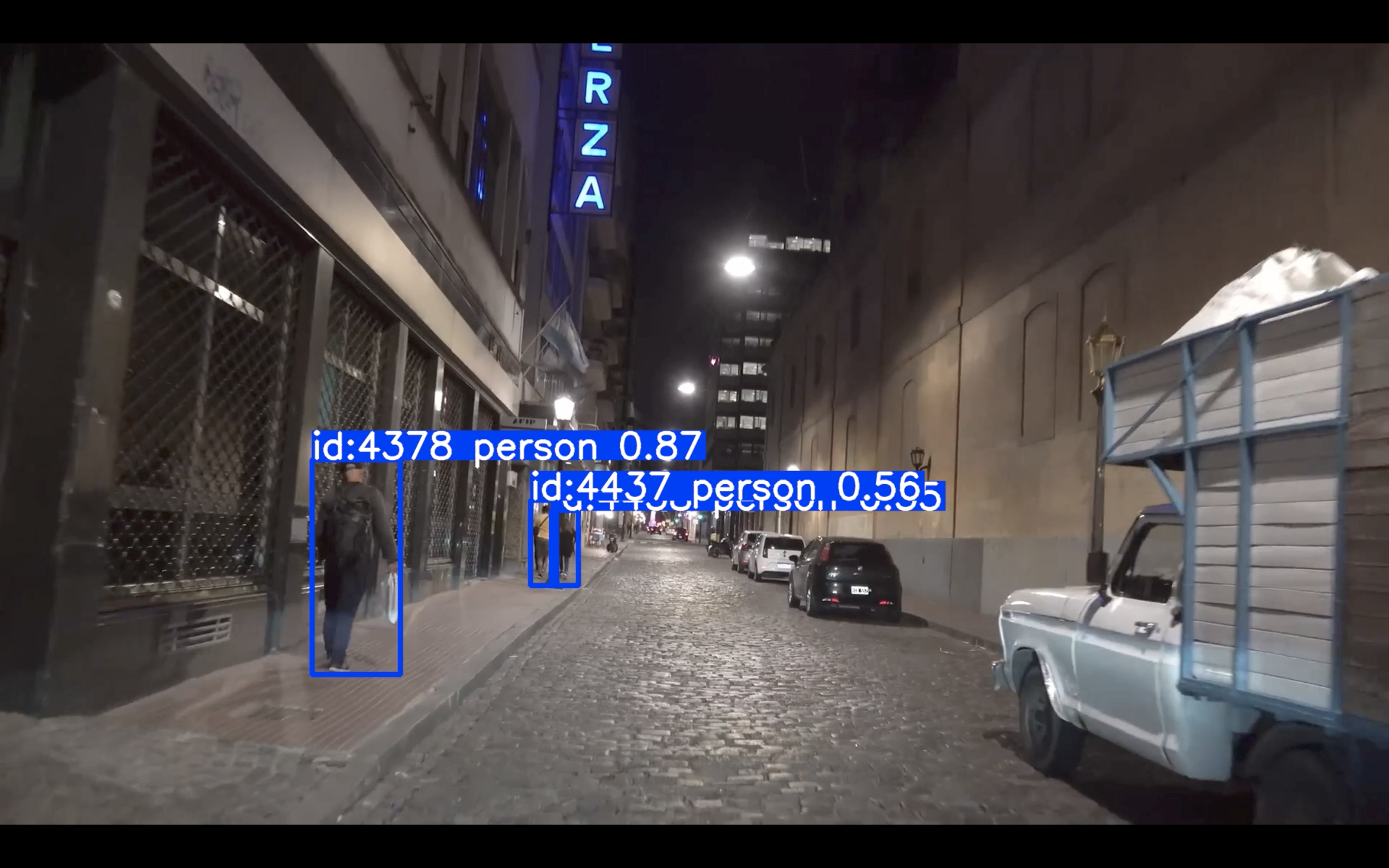}\\[-3pt]
    {\scriptsize (f)}
\end{minipage}
\hfill
\begin{minipage}[t]{0.205\textwidth}
    \centering
    \includegraphics[width=\linewidth]{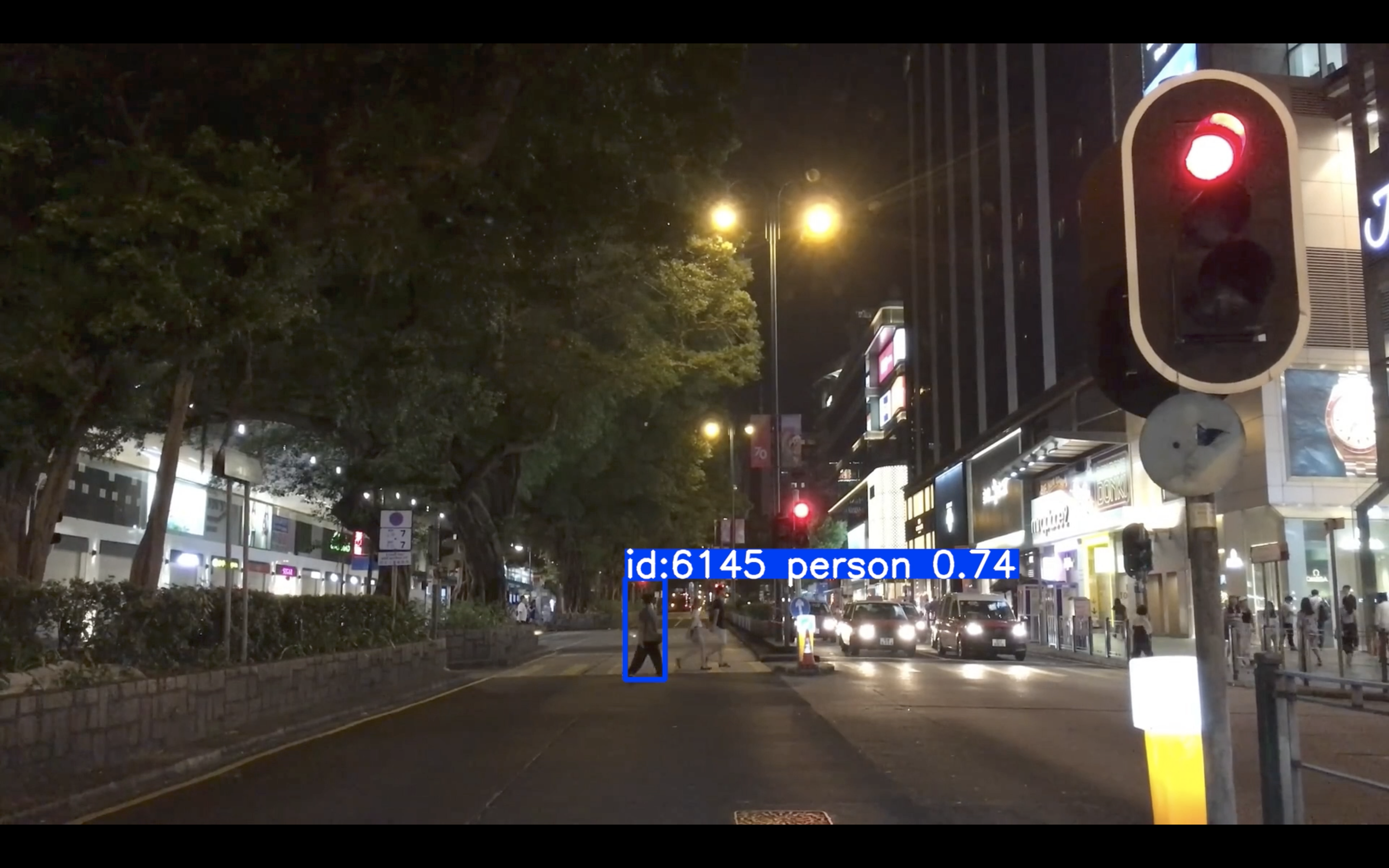}\\[-3pt]
    {\scriptsize (g)}
\end{minipage}
\hfill
\begin{minipage}[t]{0.205\textwidth}
    \centering
    \includegraphics[width=\linewidth]{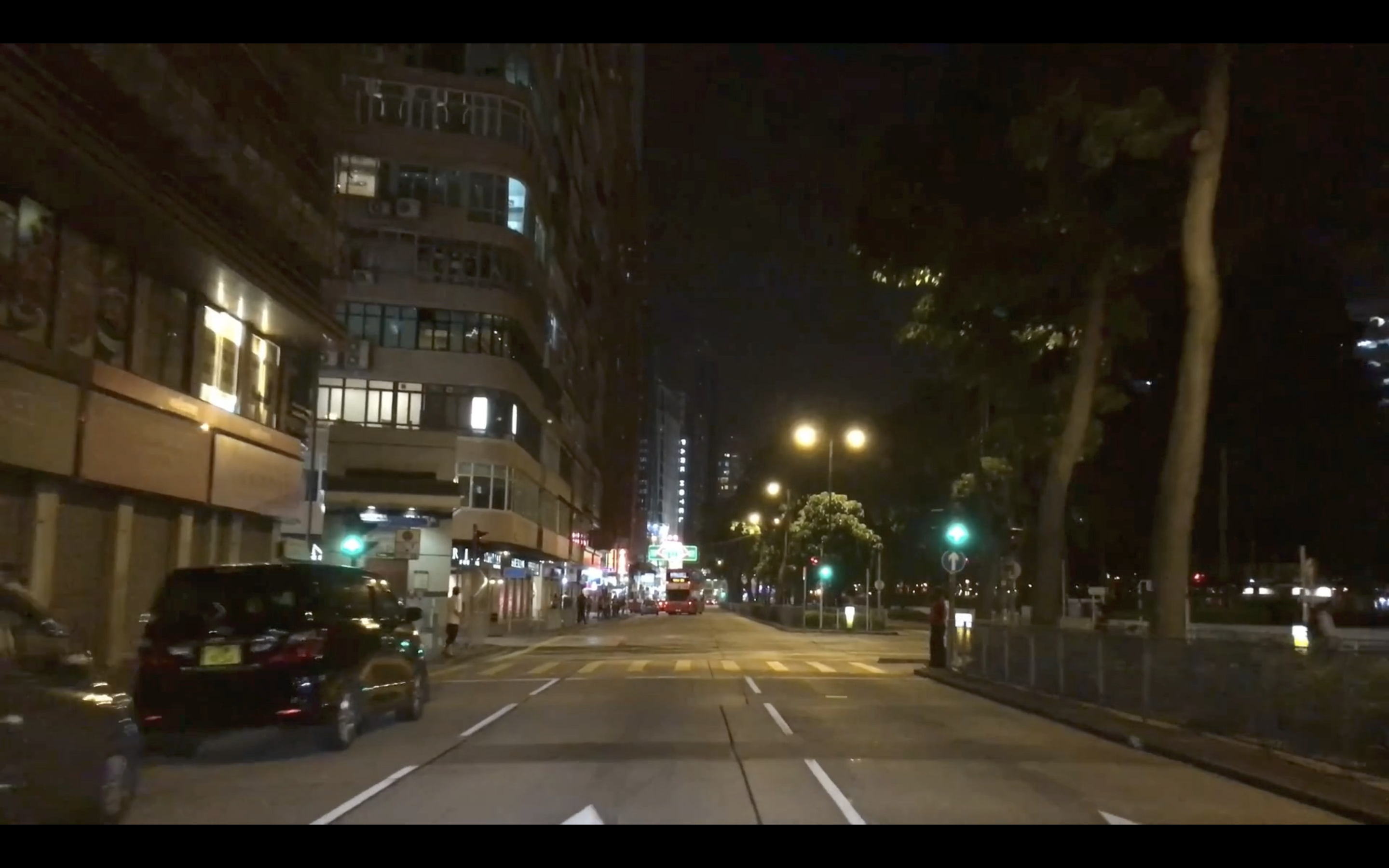}\\[-3pt]
    {\scriptsize (h)}
\end{minipage}

\vspace{2pt}


\begin{minipage}[t]{0.205\textwidth}
    \centering
    \includegraphics[width=\linewidth]{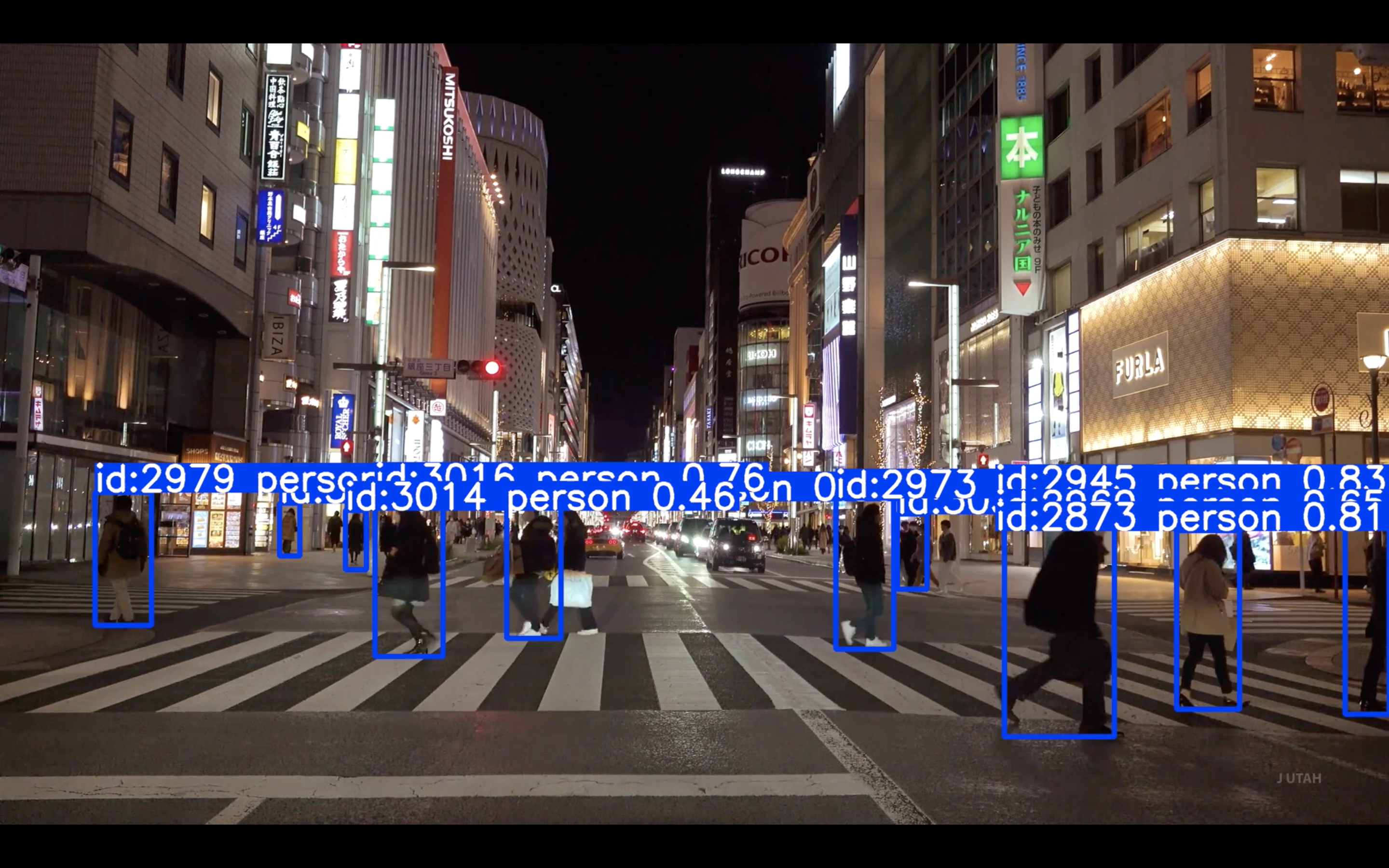}\\[-3pt]
    {\scriptsize (i)}
\end{minipage}
\hfill
\begin{minipage}[t]{0.205\textwidth}
    \centering
    \includegraphics[width=\linewidth]{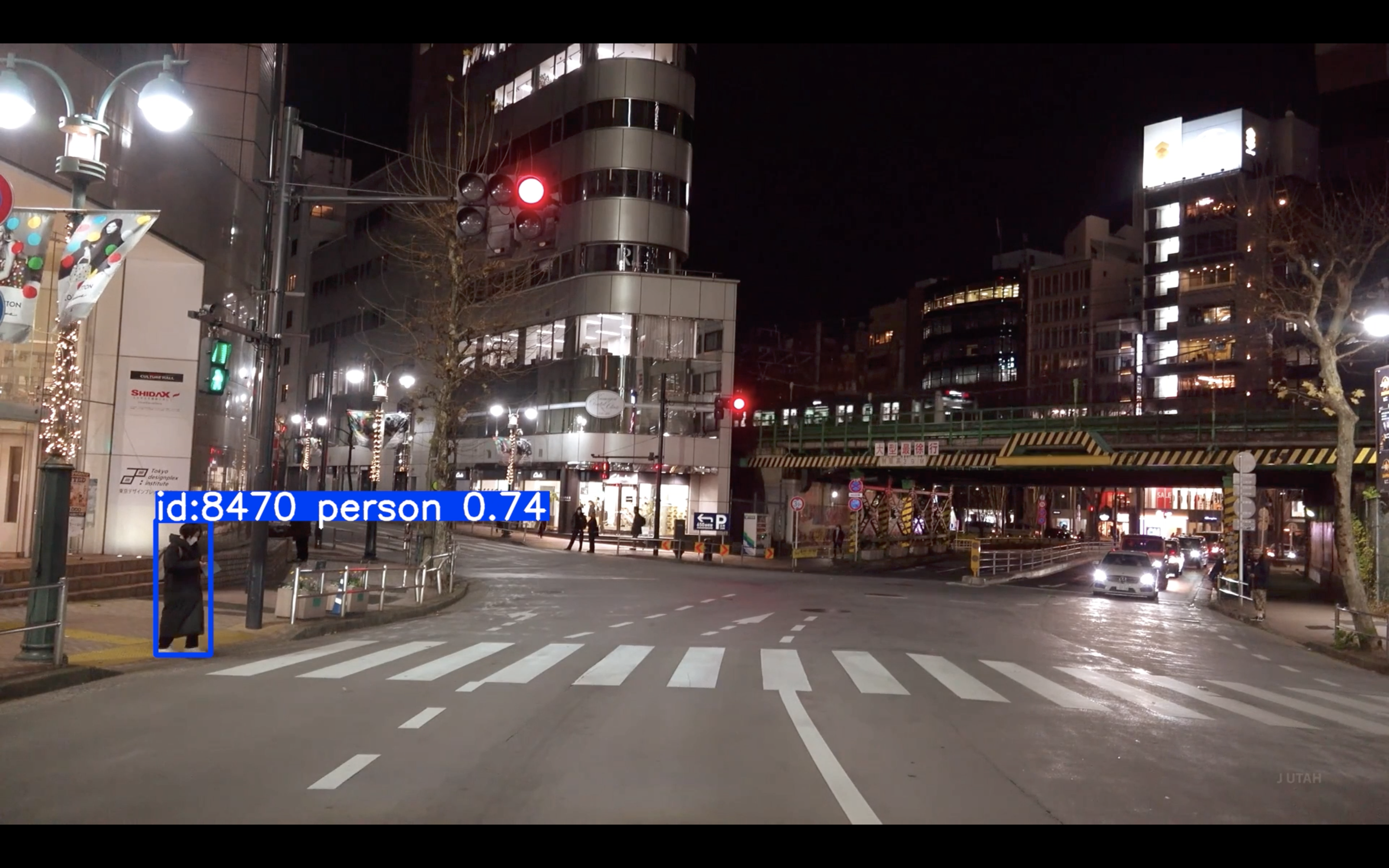}\\[-3pt]
    {\scriptsize (j)}
\end{minipage}
\hfill
\begin{minipage}[t]{0.205\textwidth}
    \centering
    \includegraphics[width=\linewidth]{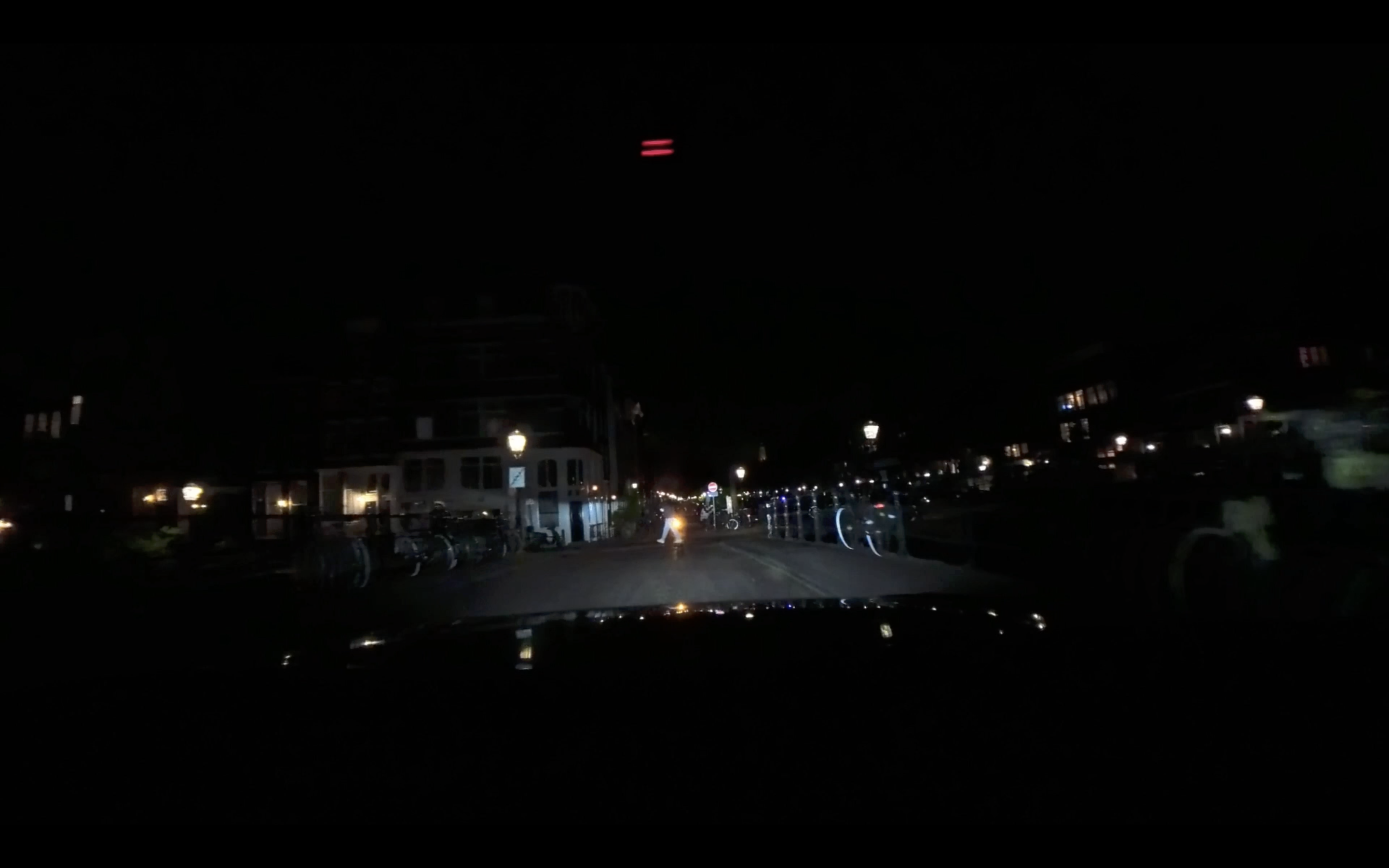}\\[-3pt]
    {\scriptsize (k)}
\end{minipage}
\hfill
\begin{minipage}[t]{0.205\textwidth}
    \centering
    \includegraphics[width=\linewidth]{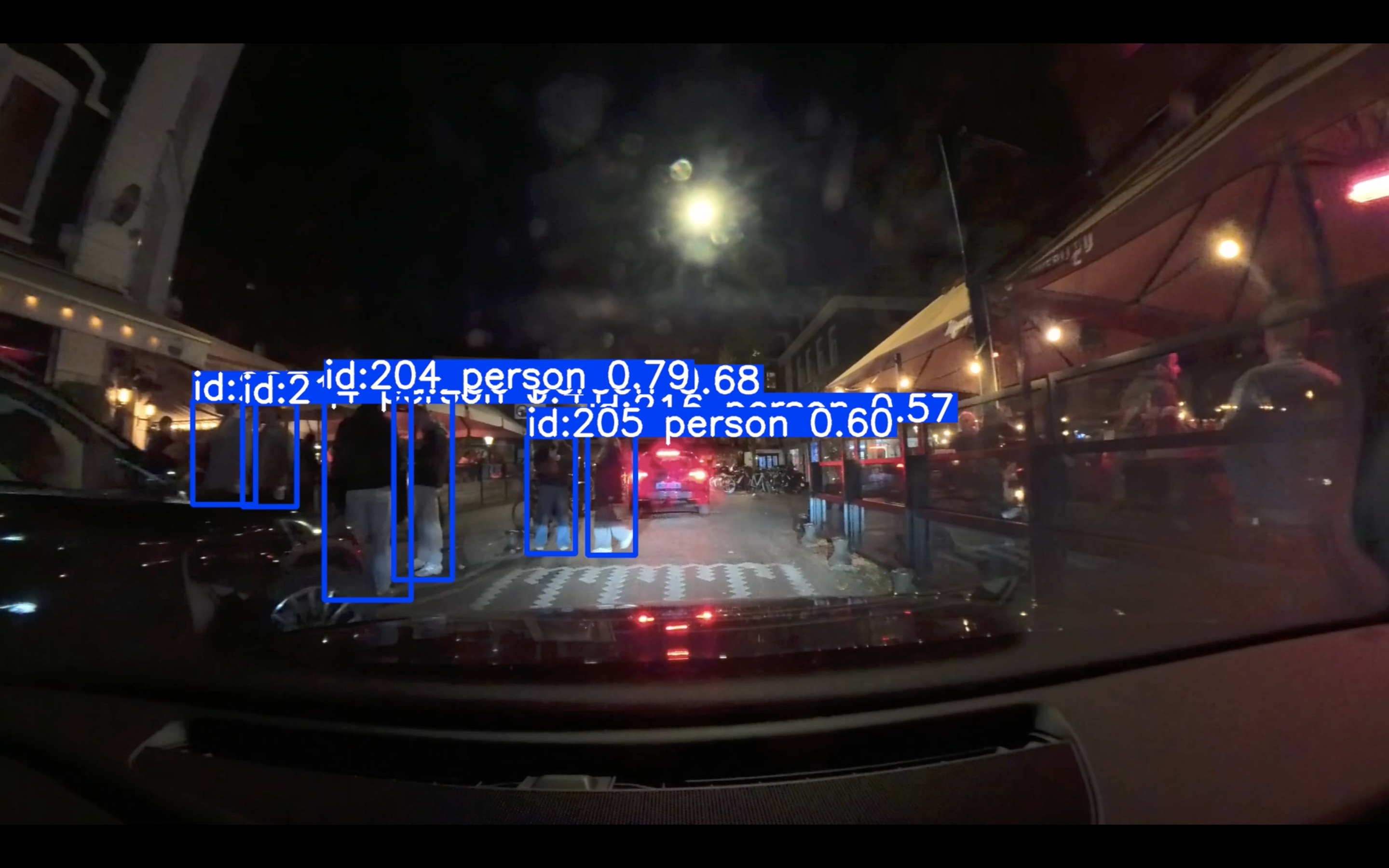}\\[-3pt]
    {\scriptsize (l)}
\end{minipage}

\vspace{1pt}

\captionof{figure}{
Visual diversity across the six cities, containing two images per city
}
\label{fig:city_diversity}

\end{strip}

\section{Benchmark}

The aim of the dataset construction is to benchmark PCIP models to gain insight into their performance under night-time conditions. In this section, the usefulness of the NightSteps dataset is demonstrated through evaluations of the AI models Gemini and ChatGPT, as well as human assessors. Results show that human assessors (accuracy: 88.17\%) perform better than Gemini (accuracy: 60\%) and ChatGPT (accuracy: 79.13\%). The differences in performance between the evaluated models demonstrate that the dataset can be used to distinguish the performance of different models and therefore can be used for benchmarking PCIP models under night-time conditions.

\subsection{Experimental Setting}
\subsubsection{Dataset and Benchmark}

The self-generated data is intended to be used in future research to evaluate the robustness of models against a baseline benchmark. In this benchmark, ground-truth crossing labels are compared with the intention predictions generated by the model. The ground-truth labels are the initial labels containing the correct information about the pedestrians’ crossing behaviour, derived from the collected data. 

In this study, the usefulness of the NightSteps dataset is demonstrated by comparing these labels with predictions from Gemini \parencite{gemini2023team}, ChatGPT \parencite{openai2026chatgpt}, and with predictions made by human participants. For this study, Gemini's 3.1 Pro extended thinking model and ChatGPT's 5.5 average thinking model were used. The aim of this benchmark is to demonstrate the usefulness of the constructed NightSteps dataset so that future research can use this dataset and results to evaluate and compare models.

\subsubsection{Experiment Settings}

From the initial data, twenty labels were selected from each video to ensure that the distribution across the six clips remains balanced and that sufficient diversity from different cities is utilised. The data recorded in Amsterdam contains only seventeen usable labels. A total of fifteen clips from Amsterdam were selected for the evaluation, resulting in a total of 115 clips to be evaluated. This is a smaller sample than the complete dataset, taking into account a realistic evaluation process and the time required for human participants. Exactly the same 115 clips are used for both the AI models and the human assessment to compare results.

The video footage of each corresponding city is edited into a sequence of individual clips. Before a clip starts, a five-second black screen shows the 'track\_id' of the pedestrian to be assessed. This is followed by a two-second clip of a pedestrian with their corresponding 'track\_id'. These two seconds represent the period immediately before the moment a pedestrian does or does not cross. As each video consists twenty usable clips of pedestrians in the specific city, the total duration of the video is approximately 145 seconds: $(2+5)*20+5$, as the video ends with a five-second black screen as well. A prediction is made based on this period leading up to the event. This prediction is labelled in the same way as the ground-truth labels: 'yes' (1) for crossing or 'no' (0) for not crossing. 

For testing with Gemini and ChatGPT, only one chat and one prompt are used for all of the clips, so there is no variation between the different clips. The output consists solely of the predicted action with 'yes' (1) for crossing or 'no' (0) for not crossing.

In addition, the 115 clips are assessed by five human participants. Before the experiment begins, participants were asked whether they hold a driving licence. Holding a driving licence may influence predictions due to driving experience. Any differences in predictions between participants with and without a driving licence can be analysed. The inputs in this human assessment experiment are the presentation of the 115 clips and the question of whether they hold a driving licence. The output is a list of the predicted labels for the 115 clips.

The images below recreate a clip of a pedestrian with 'track\_id' 505. The screen blacks out at the moment the pedestrian reaches the point where they may cross the road. The images are arranged in horizontal chronological order.

\begin{figure}[H]
    \centering

    \begin{minipage}[t]{0.48\linewidth}
        \centering
        \includegraphics[width=\linewidth]{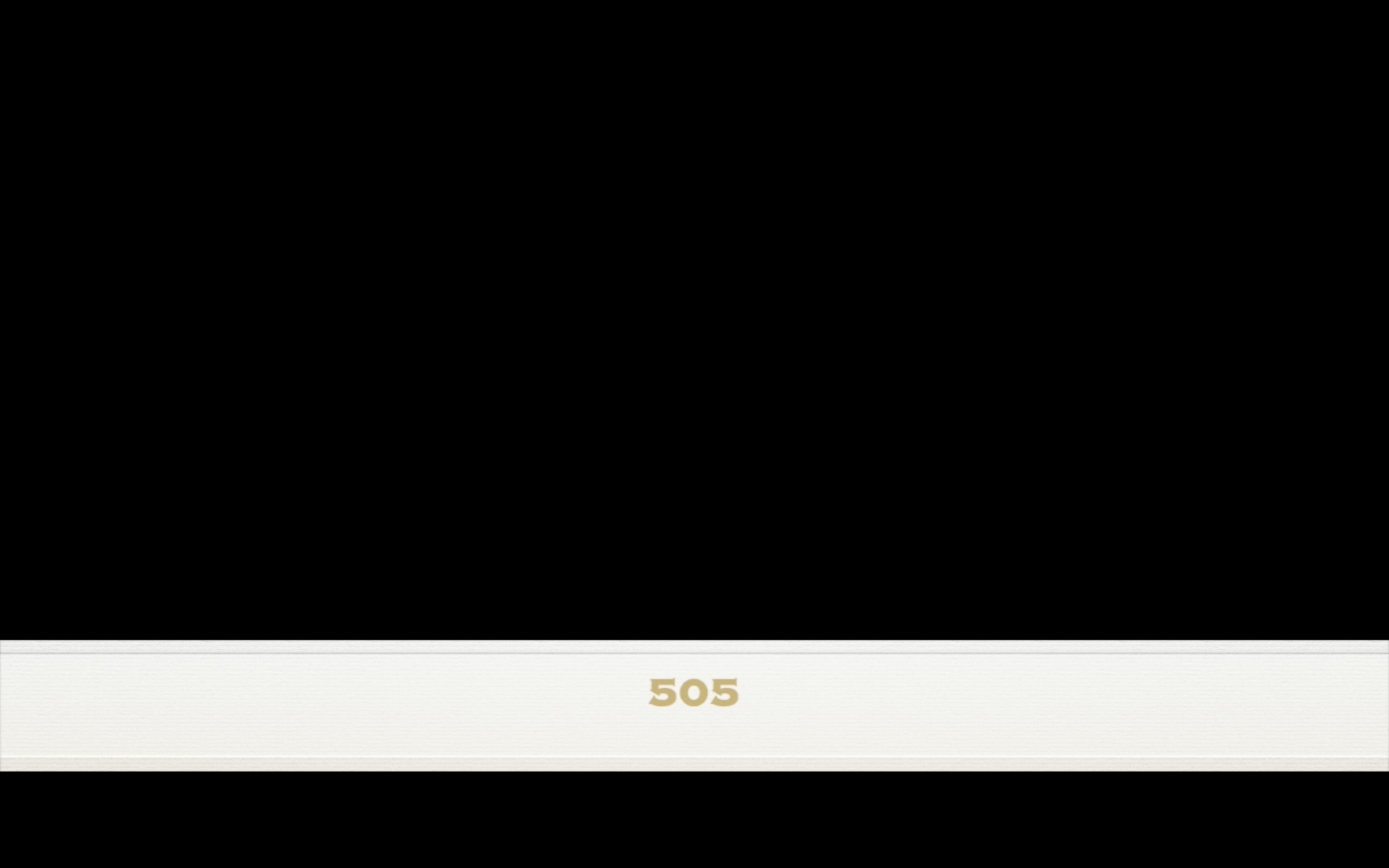}
    \end{minipage}
    \hfill
    \begin{minipage}[t]{0.48\linewidth}
        \centering
        \includegraphics[width=\linewidth]{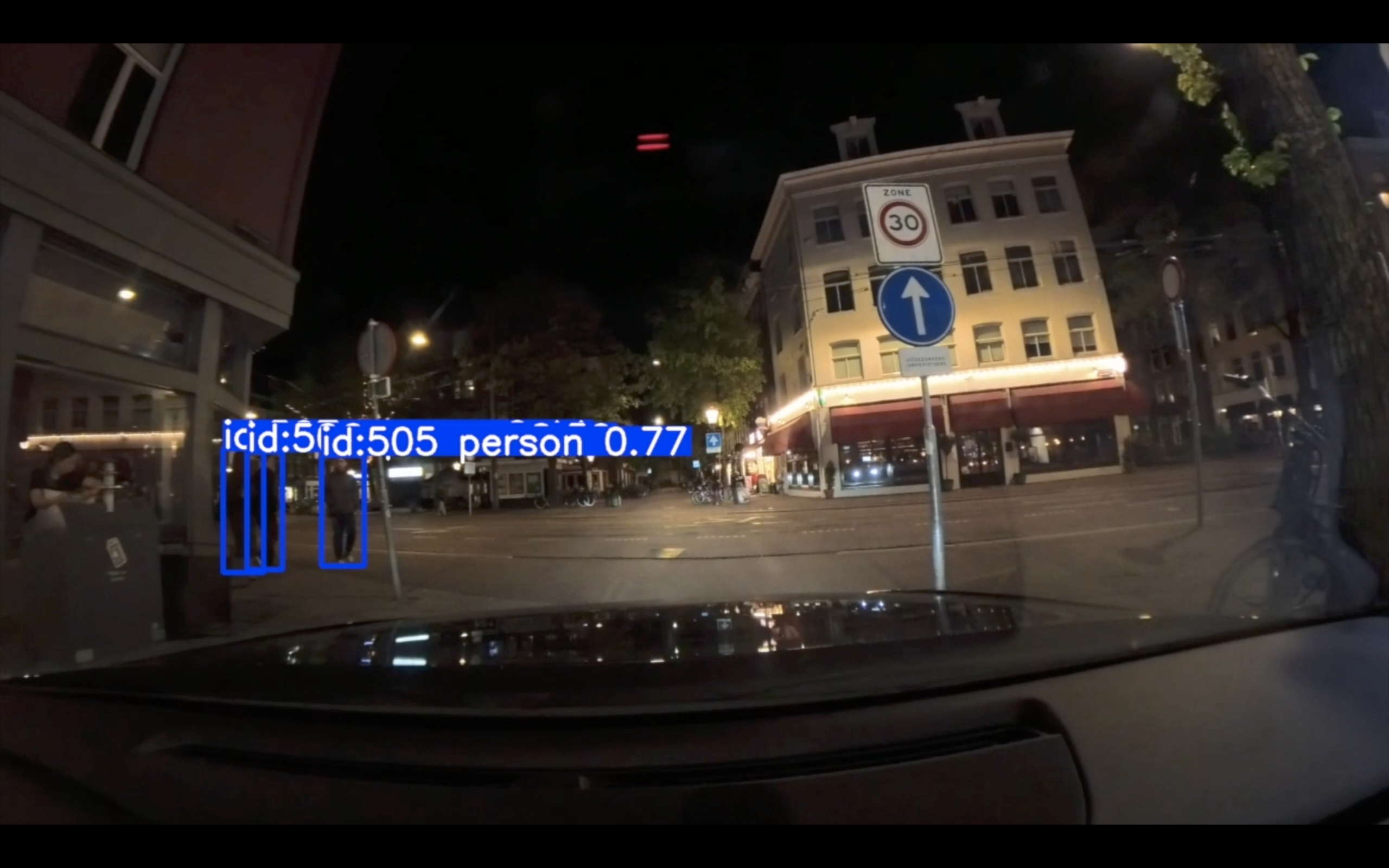}
    \end{minipage}

    \vspace{0.15cm}

    \begin{minipage}[t]{0.48\linewidth}
        \centering
        \includegraphics[width=\linewidth]{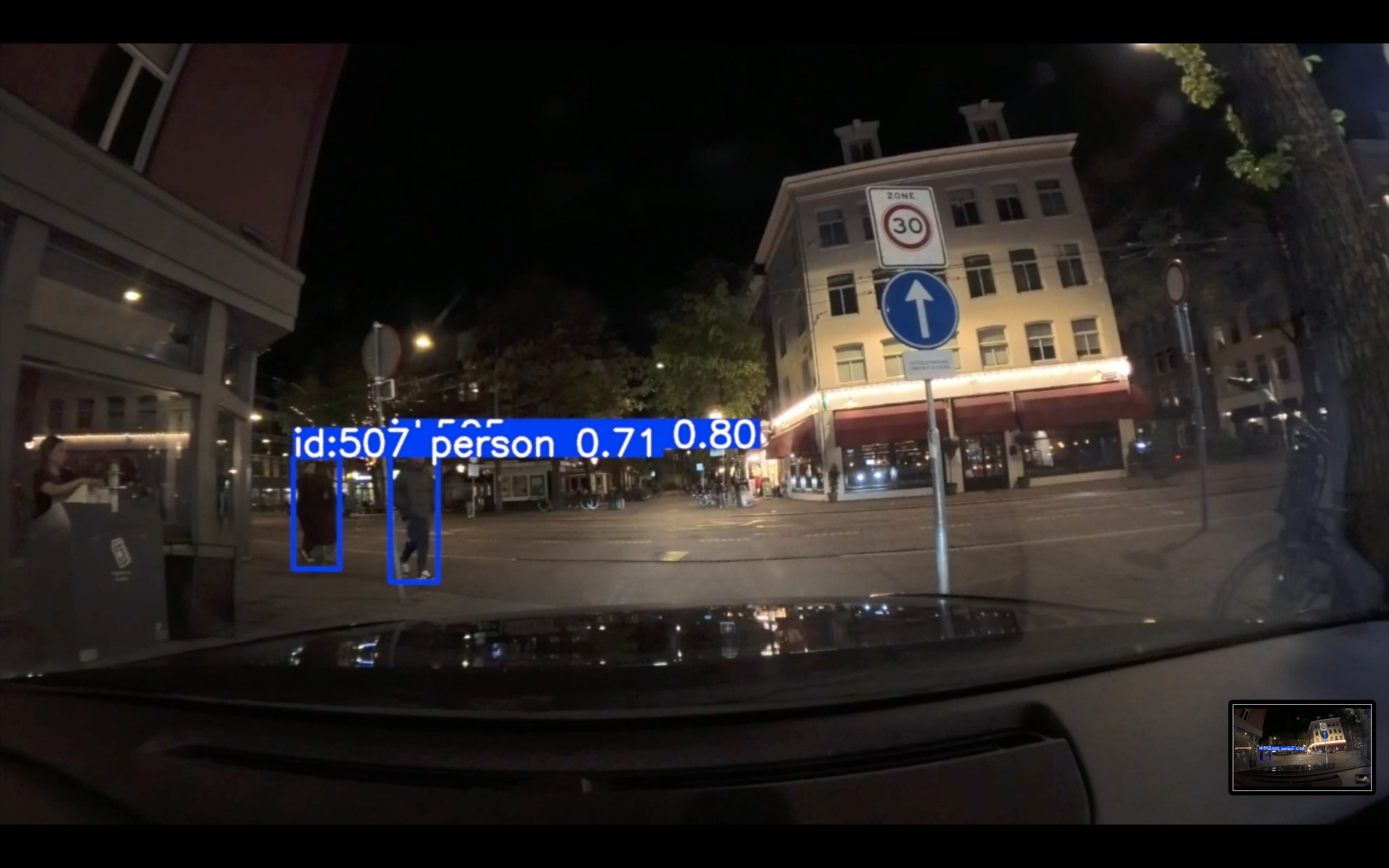}
    \end{minipage}
    \hfill
    \begin{minipage}[t]{0.48\linewidth}
        \centering
        \includegraphics[width=\linewidth]{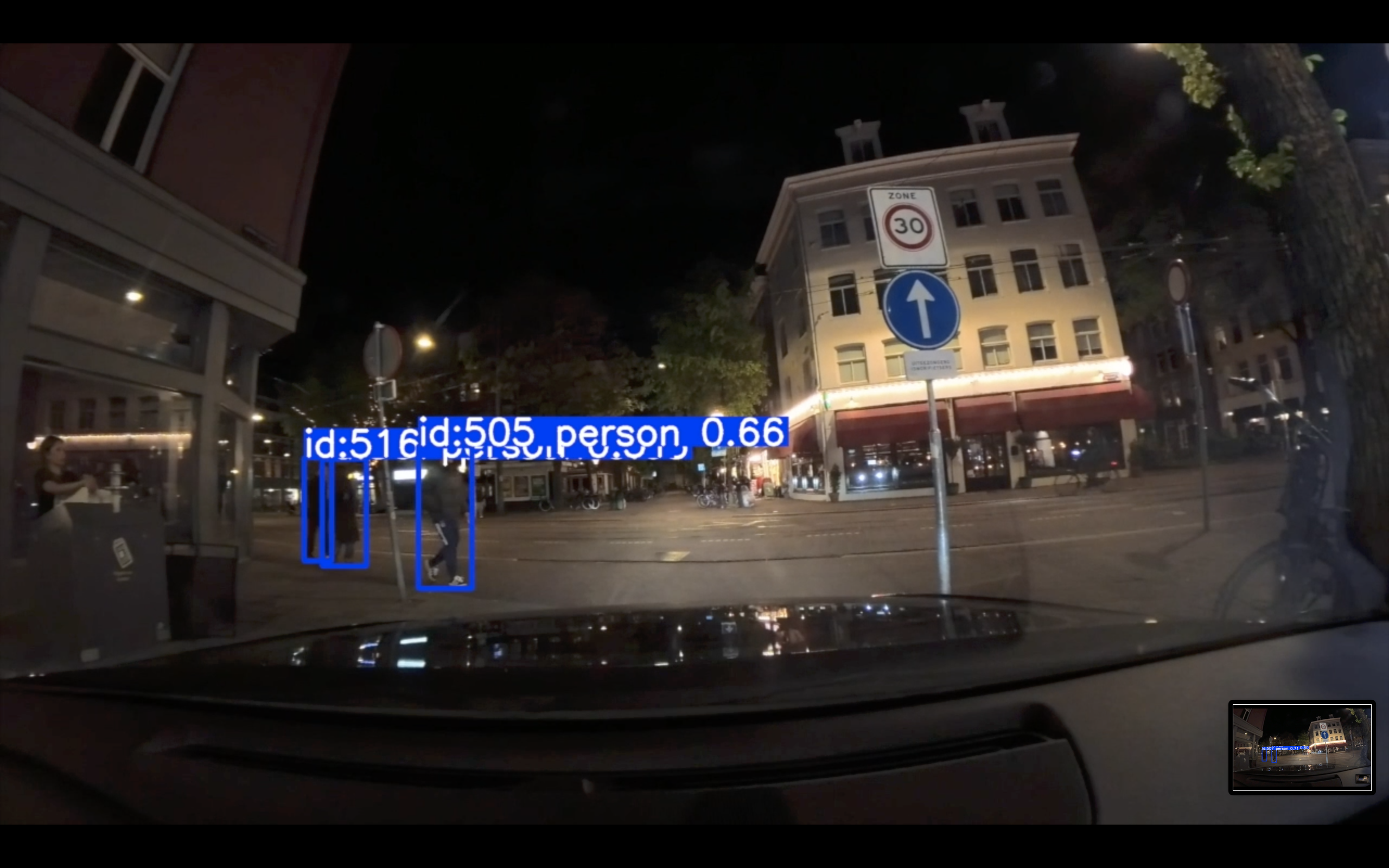}
    \end{minipage}

    \caption{Reconstructed video sequence for 'track\_id' 505}
    \label{fig:prediction_sequence}
\end{figure}

\subsubsection{Metrics}

To evaluate the suitability of the dataset for both the AI models and human assessment, predictions are compared with the ground-truth labels. Various evaluation metrics are used for this purpose: accuracy, precision, recall and an F1-score. These evaluation metrics provide insight into the predictions and are defined as:

\begin{itemize}
    \item Accuracy: gives the number of correctly predicted labels relative to the total number of predictions and indicates the overall correctness of the model.
    
    \item Precision: gives the ratio between the number of correctly predicted crossings (True positives) and the total number of clips predicted as crossings (True positives + False positives).
    
    \item Recall: gives the ratio of the number of correctly predicted crossings (True positives) and the total number of actual crossings in the dataset (True positives + False Negatives).
    
    \item F1-score: gives the harmonic mean of precision and recall to provide a balanced view of both precision and recall.
\end{itemize}

\begin{figure}[H]
    \centering
    \includegraphics[width=0.75\linewidth]{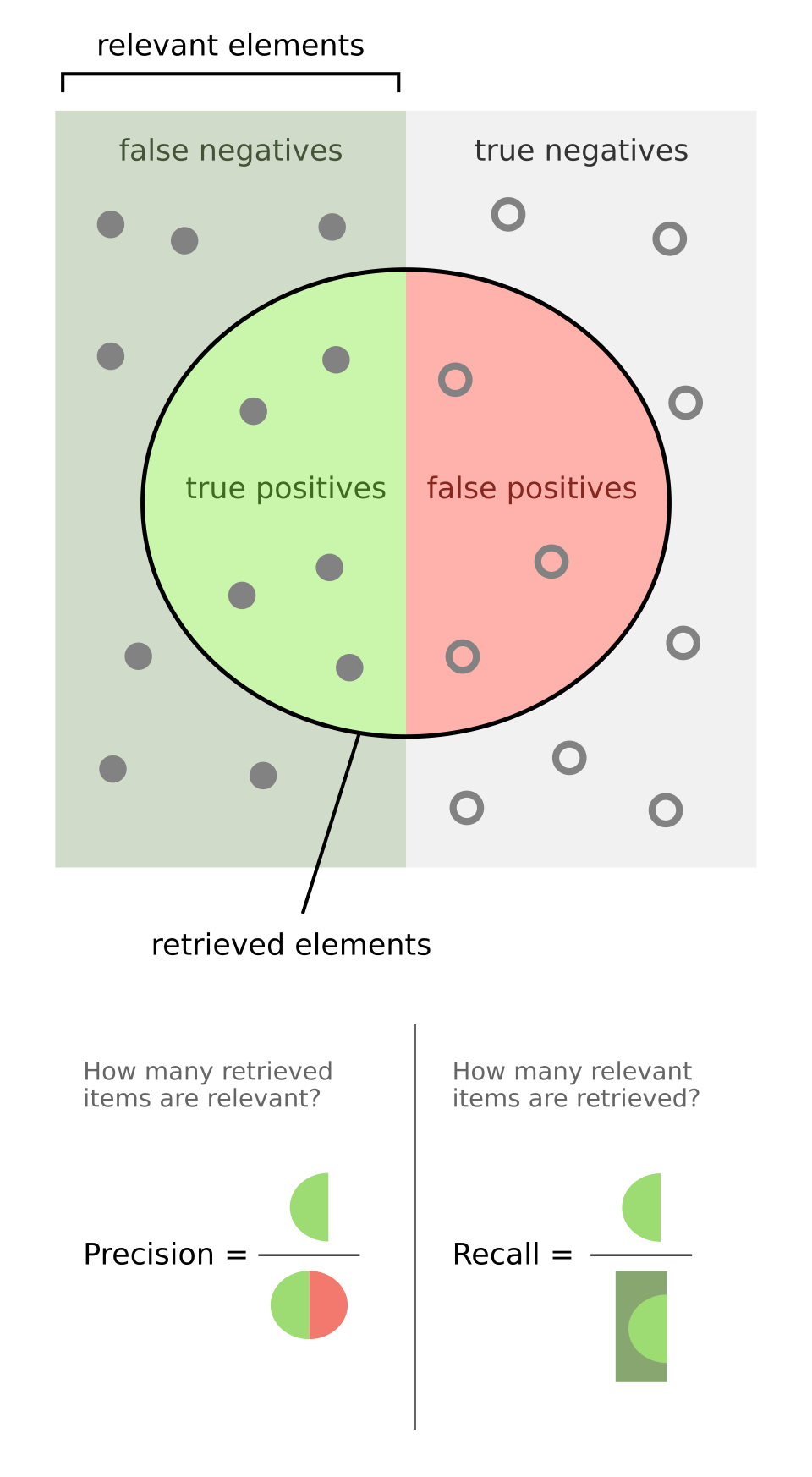}
    \caption{Visualisation of precision and recall, reproduced from \textcite{walber2014}}
    \label{fig:evaluation_metrics}
\end{figure}

\subsection{Quantitative Results}

This section presents the quantitative results of the evaluation. The performance of the Gemini model, the ChatGPT model, and the human assessment are first presented separately, followed by an overall comparison of all the clips. The results are based on the tables containing predictions and ground-truth labels.

\subsubsection{Performance Gemini}

Gemini’s performance has been evaluated using footage from the six different sub-datasets containing different tracks in different cities. The predictions regarding pedestrian crossing behaviour across these different cities are evaluated using accuracy, precision, recall and the F1-score. The table below shows both the key results per city and the overall scores.

\begin{table}[H]
    \centering
    \caption{Gemini's performance metrics per city and overall}
    \label{tab:gemini_performance}

    \resizebox{\linewidth}{!}{%
        \begin{tabular}{lccccc}
            \toprule
            \textbf{City} &
            \textbf{No.} &
            \textbf{Accuracy} &
            \textbf{Precision} &
            \textbf{Recall} &
            \textbf{F1-score} \\
            \midrule
            Amsterdam       & 15 & 46.67\% & 16.67\% & 25.00\% & 20.00\% \\
            Buenos Aires    & 20 & 50.00\% & 33.33\% & 42.86\% & 37.50\% \\
            Hong Kong       & 20 & 70.00\% & 62.50\% & 62.50\% & 62.50\% \\
            Madrid          & 20 & 60.00\% & 28.57\% & 40.00\% & 33.33\% \\
            New York City   & 20 & 70.00\% & 37.50\% & 75.00\% & 50.00\% \\
            Tokyo           & 20 & 60.00\% & 30.00\% & 75.00\% & 42.86\% \\
            \midrule
            \textbf{Overall} &
            \textbf{115} &
            \textbf{60.00\%} &
            \textbf{35.42\%} &
            \textbf{53.13\%} &
            \textbf{42.50\%} \\
            \bottomrule
        \end{tabular}%
    }
\end{table}

Across the entire dataset, Gemini correctly predicted 69 out of 115 potential crossings, resulting in an accuracy of 60\%. The overall precision is 35.42\%, which represents the number of correctly predicted crossings relative to the total number of predicted crossings (17/48). The total recall is 53.13\%, which represents the number of correctly predicted crossings relative to the actual number of crossings (17/32). Together, these scores give a total F1-score of 42.50\%.

The predictions show that the highest accuracy is achieved in videos in both Hong Kong and New York City, with an accuracy of 70\%. The lowest accuracy is achieved in videos in Amsterdam with a score of 46.67\%. The highest number of correctly predicted crossings achieved by the Gemini model, are both videos from New York City and Tokyo, with a recall of 75\%.

\subsubsection{Performance ChatGPT}

ChatGPT’s performance has been evaluated on the same six videos. The predictions regarding pedestrian crossing behaviour across these different cities are also evaluated using accuracy, precision, recall and the F1-score. The table below shows both the key results per city and the overall scores.

\begin{table}[H]
    \centering
    \caption{ChatGPT's performance metrics per city and overall}
    \label{tab:chatgpt_performance}

    \resizebox{\linewidth}{!}{%
        \begin{tabular}{lccccc}
            \toprule
            \textbf{City} &
            \textbf{No.} &
            \textbf{Accuracy} &
            \textbf{Precision} &
            \textbf{Recall} &
            \textbf{F1-score} \\
            \midrule
            Amsterdam       & 15 & 73.33\%  & 50.00\%  & 50.00\%  & 50.00\% \\
            Buenos Aires    & 20 & 65.00\%  & 50.00\%  & 71.43\%  & 58.82\% \\
            Hong Kong       & 20 & 70.00\%  & 66.67\%  & 50.00\%  & 57.14\% \\
            Madrid          & 20 & 80.00\%  & 66.67\%  & 40.00\%  & 50.00\% \\
            New York City   & 20 & 85.00\%  & 57.14\%  & 100.00\% & 72.73\% \\
            Tokyo           & 20 & 100.00\% & 100.00\% & 100.00\% & 100.00\% \\
            \midrule
            \textbf{Overall} &
            \textbf{115} &
            \textbf{79.13\%} &
            \textbf{61.76\%} &
            \textbf{65.63\%} &
            \textbf{63.64\%} \\
            \bottomrule
        \end{tabular}%
    }
\end{table}

Across the entire dataset, ChatGPT correctly predicted 91 out of 115 potential crossings. This equates to an accuracy of 79.13\%. The number of correctly predicted crossings relative to the total number of predicted crossings is 21 out of 34, which equates to a precision of 61.76\%. The actual number of crossings is 32, resulting in a recall of 65.63\%. Combined, these scores give a total F1-score of 63.64\%.

Here too, the accuracy of the predictions in a particular city stands out, namely with an accuracy of 100\% in Tokyo. In this city, four people crossed the road, which ChatGPT all predicted correctly. Furthermore, no other crossings were predicted that did not contain a crossing label. This means that all situations were predicted correctly and all scores come out at 100\%. The lowest accuracy was achieved in Buenos Aires, with an accuracy of 65\%. However, the recall here is above average. In addition to the F1-score of 100\% in Tokyo, New York City also achieved relatively high scores with an F1-score of 72.73\% and an accuracy of 85\%.

\subsubsection{Performance Human Assessment}

The human assessment was also carried out using video footage from the six different cities. The participants have stated in advance that they all hold a driving licence, which implies an understanding of road traffic. Differences between participants regarding driving licence ownership cannot be ruled out. The results of the human-predicted crossings, along with the corresponding scores, are shown in the table below.

\begin{table}[H]
    \centering
    \caption{Human performance metrics per city and overall}
    \label{tab:human_performance}

    \resizebox{\linewidth}{!}{%
        \begin{tabular}{lccccc}
            \toprule
            \textbf{City} &
            \textbf{No.} &
            \textbf{Accuracy} &
            \textbf{Precision} &
            \textbf{Recall} &
            \textbf{F1-score} \\
            \midrule
            Amsterdam       & 15 & 90.66\% & 90.00\% & 75.00\%  & 81.43\% \\
            Buenos Aires    & 20 & 80.00\% & 65.67\% & 91.43\%  & 76.21\% \\
            Hong Kong       & 20 & 91.00\% & 86.20\% & 92.50\%  & 88.93\% \\
            Madrid          & 20 & 89.00\% & 82.00\% & 72.00\%  & 76.45\% \\
            New York City   & 20 & 92.00\% & 76.67\% & 100.00\% & 85.33\% \\
            Tokyo           & 20 & 87.00\% & 69.00\% & 75.00\%  & 70.81\% \\
            \midrule
            \textbf{Overall} &
            \textbf{115} &
            \textbf{88.17\%} &
            \textbf{75.54\%} &
            \textbf{85.62\%} &
            \textbf{80.21\%} \\
            \bottomrule
        \end{tabular}%
    }
\end{table}

An overall accuracy of 88.17\% was achieved, which equates to an average of 101.4 correct predictions out of a total of 115 potential crossings. The average overall precision is 75.54\% and the average overall recall is 85.62\%. These scores result in a combined F1-score of 80.21\%.

Although the results are close, the participants are particularly good at predicting crossing behaviour in New York City, with an accuracy of 92\%. The lowest accuracy is achieved for videos in Buenos Aires, with an accuracy of 80\%. The highest number of correctly predicted
crossings achieved by human predictions is in New York City, with a recall of 100\%.

\subsubsection{Comparison of Models}

A comparison of Tables 4.1, 4.2, and 4.3 shows that human predictions outperform the Gemini and ChatGPT models in every aspect, with an accuracy of 88.17\% and an F1-score of 80.21\%. The ChatGPT model performed better than the Gemini model, with an accuracy of 79.13\% and an F1-score of 63.64\%, while Gemini achieved lower scores with an accuracy of 60\% and an F1-score of 42.50\%. There are also significant differences in prediction performance across different cities. Gemini achieves an F1-score of just 42.86\% in Tokyo, while human input and ChatGPT significantly outperform this with F1-scores of 70.81\% and even 100\%. However, it is evident that both AI models achieve their lowest F1-scores in Amsterdam, whereas humans score average here. The chart below illustrates the differences across the models and the cities according to the F1-scores. This chart clearly shows that human participants outperforms both AI models. The bars also show that ChatGPT generally scores higher than Gemini. The models are ordered according to their performance from left to right, from lowest to highest.

\begin{figure}[H]
    \centering

    \begin{tikzpicture}
        \begin{axis}[
            ybar,
            bar width=3.5pt,
            width=\linewidth,
            height=0.78\linewidth,
            ymin=0,
            ymax=110,
            ylabel={F1-score (\%)},
            symbolic x coords={
                Amsterdam,
                Buenos Aires,
                Hong Kong,
                Madrid,
                New York City,
                Tokyo
            },
            xtick=data,
            x tick label style={
                rotate=45,
                anchor=east,
                font=\scriptsize
            },
            y tick label style={
                font=\scriptsize
            },
            ylabel style={
                font=\small
            },
            legend style={
                at={(0.5,-0.32)},
                anchor=north,
                legend columns=3,
                font=\scriptsize
            },
            grid=major,
            enlarge x limits=0.08
        ]

        \addplot coordinates {
            (Amsterdam,20.00)
            (Buenos Aires,37.50)
            (Hong Kong,62.50)
            (Madrid,33.33)
            (New York City,50.00)
            (Tokyo,42.86)
        };

        \addplot coordinates {
            (Amsterdam,50.00)
            (Buenos Aires,58.82)
            (Hong Kong,57.14)
            (Madrid,50.00)
            (New York City,72.73)
            (Tokyo,100.00)
        };

        \addplot coordinates {
            (Amsterdam,81.43)
            (Buenos Aires,76.21)
            (Hong Kong,88.93)
            (Madrid,76.45)
            (New York City,85.33)
            (Tokyo,70.81)
        };

        \legend{Gemini, ChatGPT, Human}

        \end{axis}
    \end{tikzpicture}

    \caption{Comparison of F1-scores across cities for Gemini, ChatGPT, and human participants}
    \label{fig:f1_score_comparison}
\end{figure}
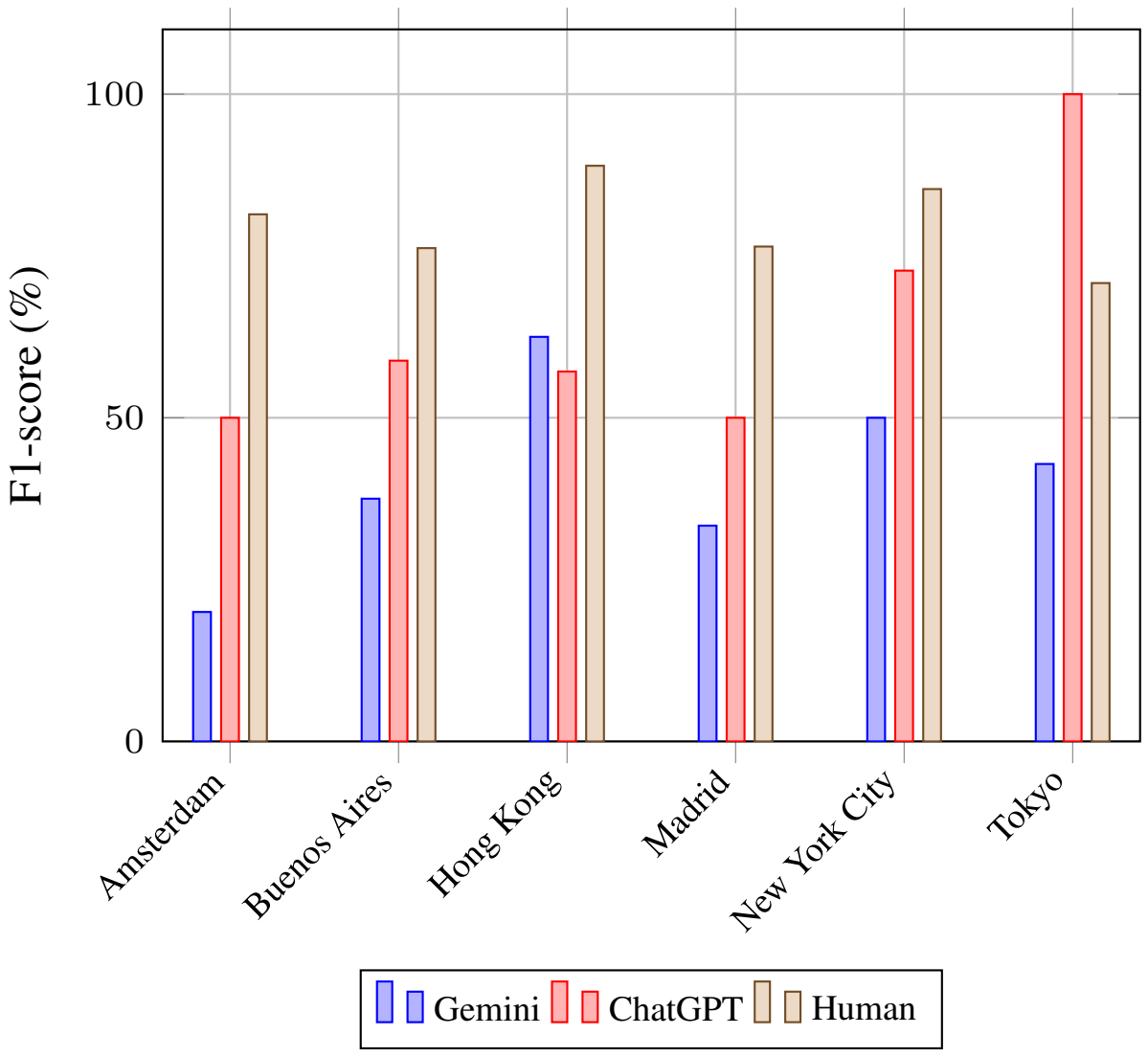

\subsection{Qualitative Results and Failure Case Study}

In addition to evaluating the models using metrics, individual clips were also assessed where the models performed either very poorly or very well. The focus here will be on the results of Gemini’s model, as it performed significantly worse than the human assessments. This section will examine various cases including incorrect predictions made by Gemini and ChatGPT, where human predictions differed from Gemini’s prediction, cases in which people collectively made an incorrect prediction, cases in which all models failed, and cases where both AI models and human predictions are correct.

Gemini's model produced the worst predictions in Amsterdam. Eight out of fifteen predictions did not match the ground-truth labels. One such case is a clip showing several people standing close together in a poorly illuminated environment. This concerns the pedestrian with 'track\_id' 249. According to the ground-truth label, the pedestrian is not crossing the road, while Gemini is convinced that the pedestrian will cross in front of the car. The grouping of people and the low illumination make this case particularly challenging for the model compared to a situation where a single pedestrian is visible in a better-lit environment. The sequence of images below illustrates the situation in this case.

\begin{figure}[H]
    \centering

    \begin{minipage}[t]{0.315\linewidth}
        \centering
        \includegraphics[width=\linewidth]{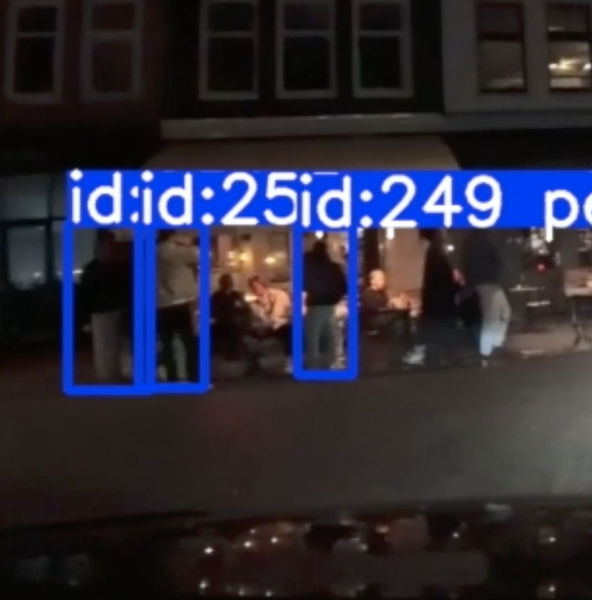}
    \end{minipage}
    \hfill
    \begin{minipage}[t]{0.315\linewidth}
        \centering
        \includegraphics[width=\linewidth]{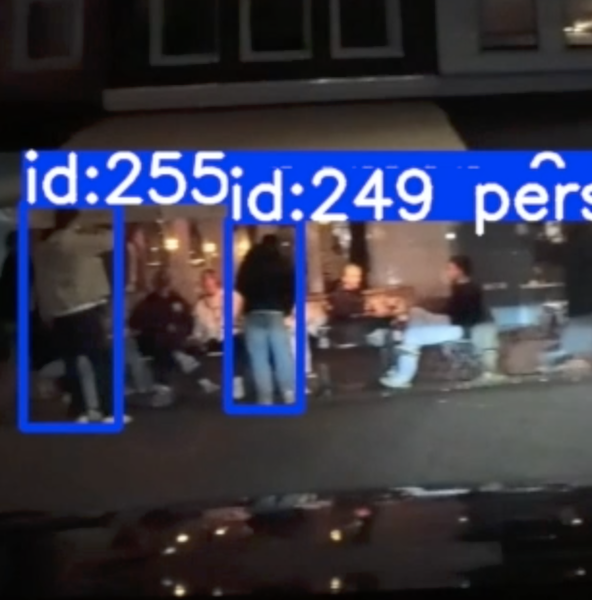}
    \end{minipage}
    \hfill
    \begin{minipage}[t]{0.315\linewidth}
        \centering
        \includegraphics[width=\linewidth]{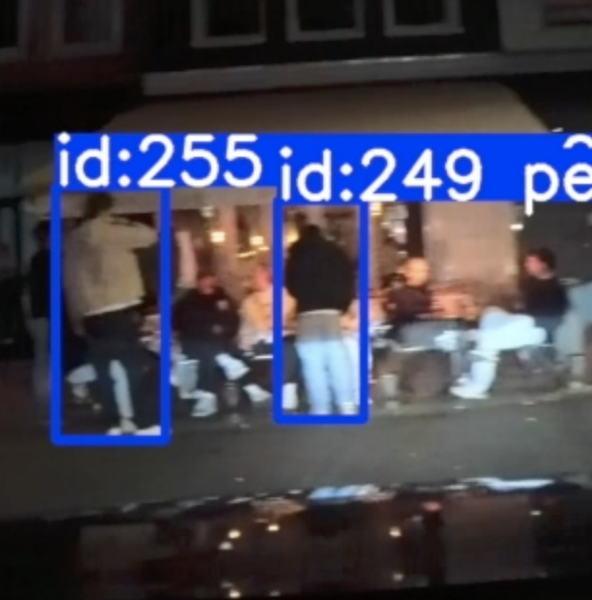}
    \end{minipage}

    \caption{Failure case in poorly illuminated group setting}
    \label{fig:failure_group_setting}
\end{figure}

A scenario ChatGPT struggles with is a clip from Hong Kong, where a vehicle is approaching at a relatively high speed, meaning the person is detected quite late. ChatGPT’s model assumes that the person with 'track\_id' 8053 is about to cross the road, whereas the ground-truth label indicates that the person remains standing at the side of the road. Because the person is detected late, the model has only a very short time to make a prediction. As a result, the model is less certain of its prediction. However, the Gemini model did make a correct prediction here, which makes the case debatable. All participants also made the correct prediction. The situation in this case is shown below.

\begin{figure}[H]
    \centering

    \begin{minipage}[t]{0.315\linewidth}
        \centering
        \includegraphics[width=\linewidth]{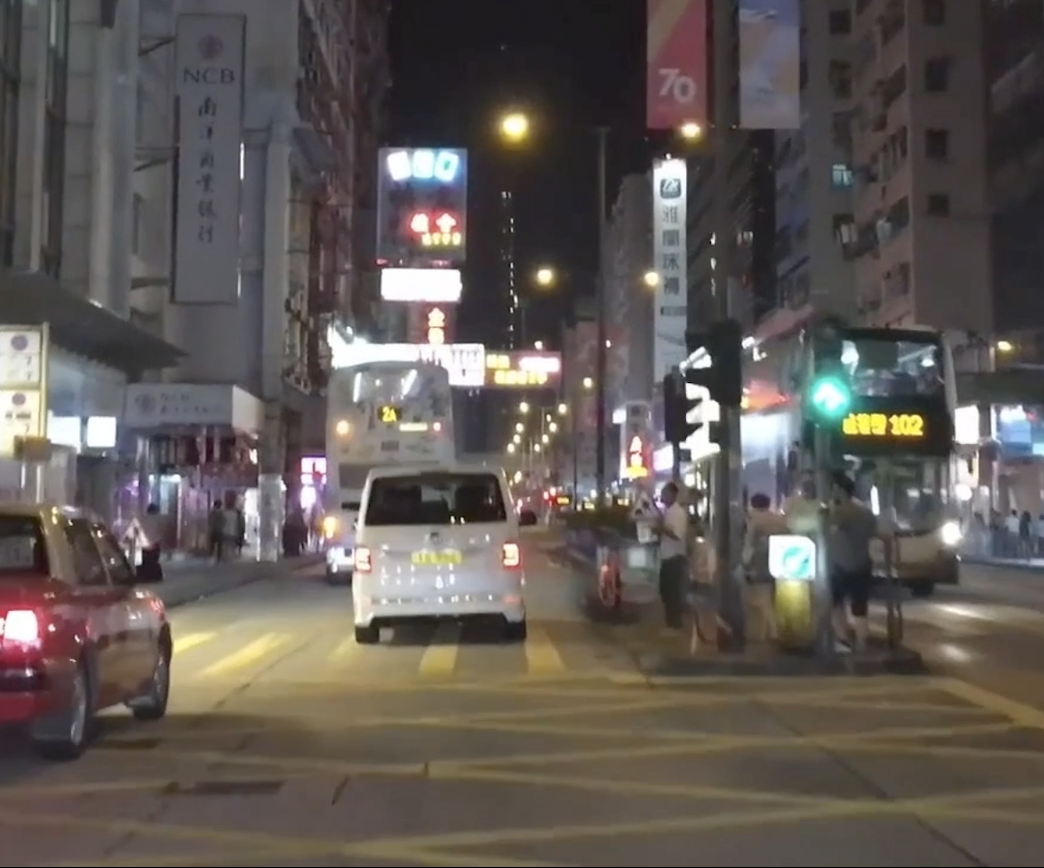}
    \end{minipage}
    \hfill
    \begin{minipage}[t]{0.315\linewidth}
        \centering
        \includegraphics[width=\linewidth]{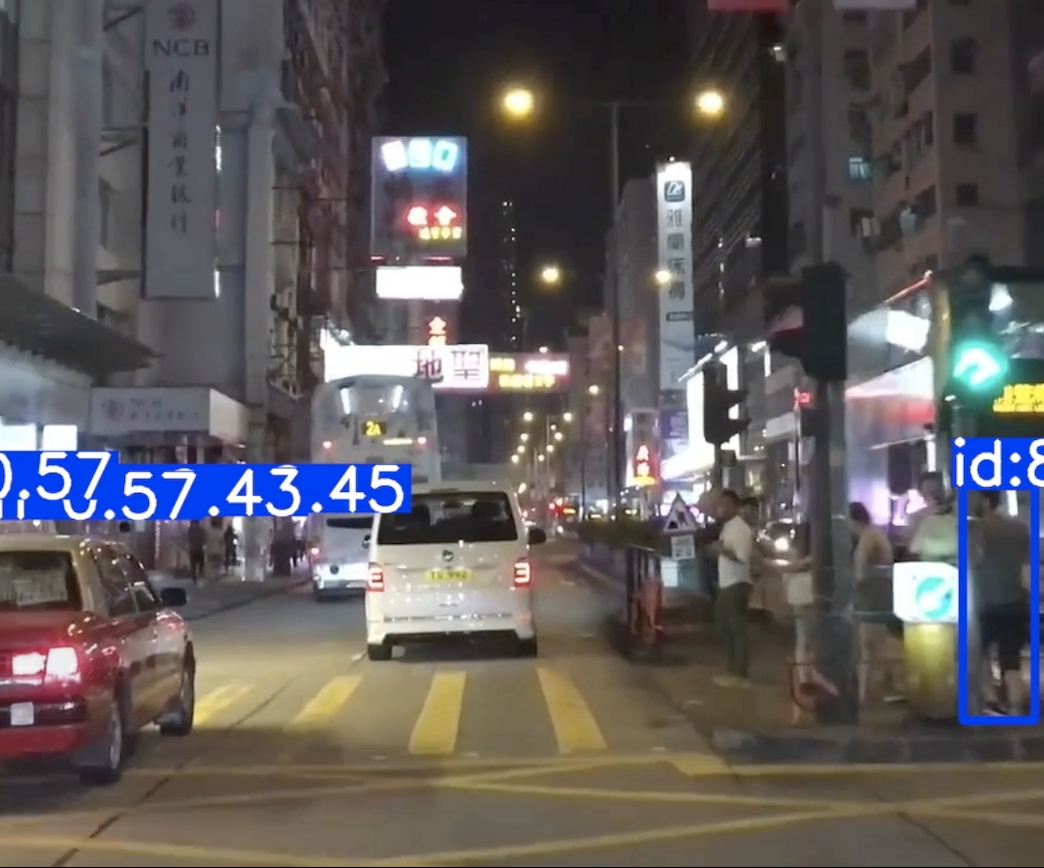}
    \end{minipage}
    \hfill
    \begin{minipage}[t]{0.315\linewidth}
        \centering
        \includegraphics[width=\linewidth]{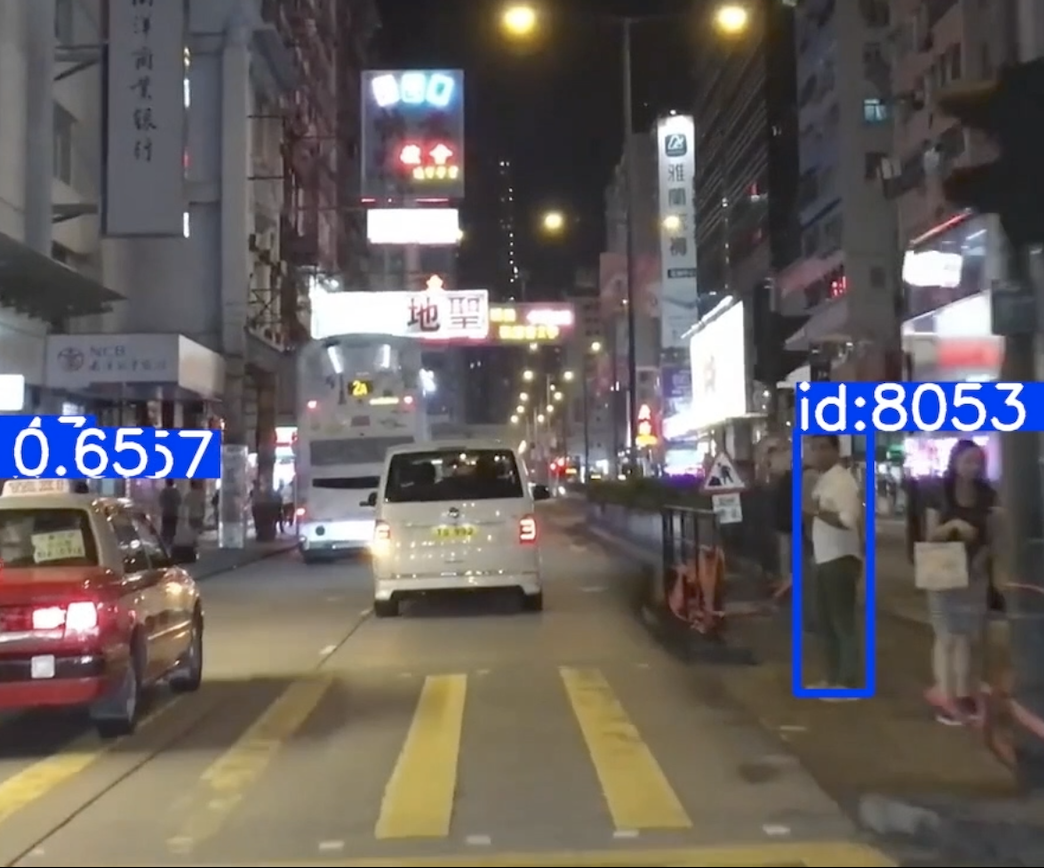}
    \end{minipage}

    \caption{Failure case with late detection}
    \label{fig:failure_late_detection}
\end{figure}

Another case from New York City shows that all five participants predict that a pedestrian is about to cross, while Gemini's model predicts that the pedestrian will not cross. The pedestrian with 'track\_id' 1039 in fact will cross. In this case too, the illumination of the pedestrian is poor, but at a certain point the pedestrian is also obscured by a lamppost. This immediately results in a gap in detection. The participants have no difficulty tracking the pedestrian despite this obstruction, while Gemini’s model is unable to make an accurate prediction. This case is also shown in the images below, where the last image clearly shows the misdetection caused by the lamppost.

\begin{figure}[H]
    \centering

    \begin{minipage}[t]{0.315\linewidth}
        \centering
        \includegraphics[width=\linewidth]{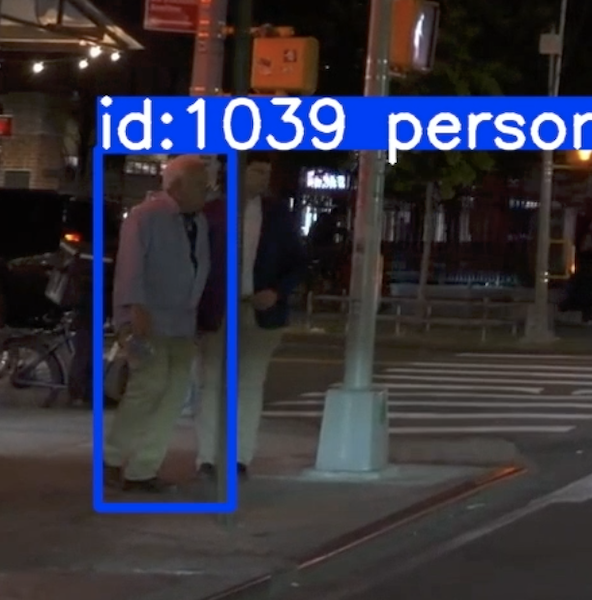}
    \end{minipage}
    \hfill
    \begin{minipage}[t]{0.315\linewidth}
        \centering
        \includegraphics[width=\linewidth]{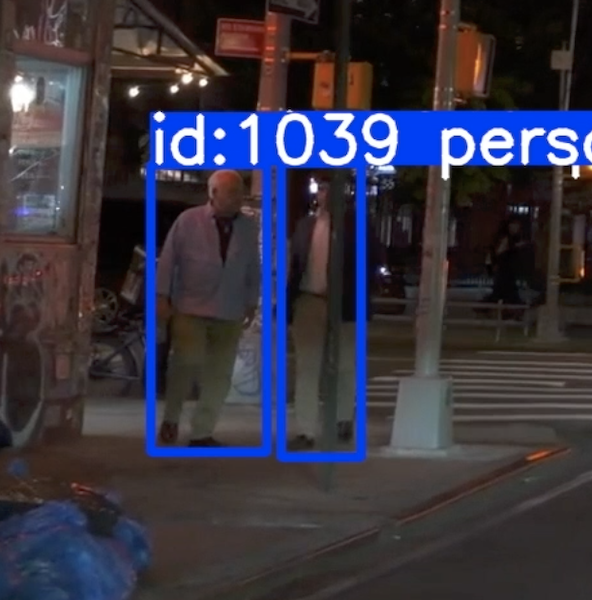}
    \end{minipage}
    \hfill
    \begin{minipage}[t]{0.315\linewidth}
        \centering
        \includegraphics[width=\linewidth]{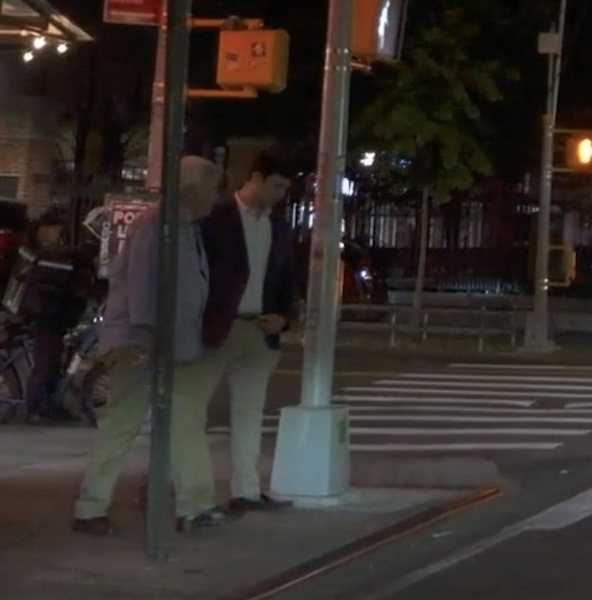}
    \end{minipage}

    \caption{Failure case with obstructed pedestrian by lamppost in poorly illuminated environment}
    \label{fig:failure_lamppost}
\end{figure}

In this case in Buenos Aires, the pedestrian with 'track\_id' 2967 is more clearly illuminated compared to previous cases. However, the pedestrian becomes visible briefly after emerging from behind a large object. Due to the high speed at which the pedestrian is approaching the road, it appears the pedestrian is about to cross. However, according to the ground-truth label, the person stops at the last moment and does not cross. This case therefore demonstrates that lighting is not always the problem, but that a pedestrian’s speed also influences the prediction of pedestrian crossing behaviour. Both AI models made a correct prediction, but all human assessors failed. The case is shown in the images below.

\begin{figure}[H]
    \centering

    \begin{minipage}[t]{0.315\linewidth}
        \centering
        \includegraphics[width=\linewidth]{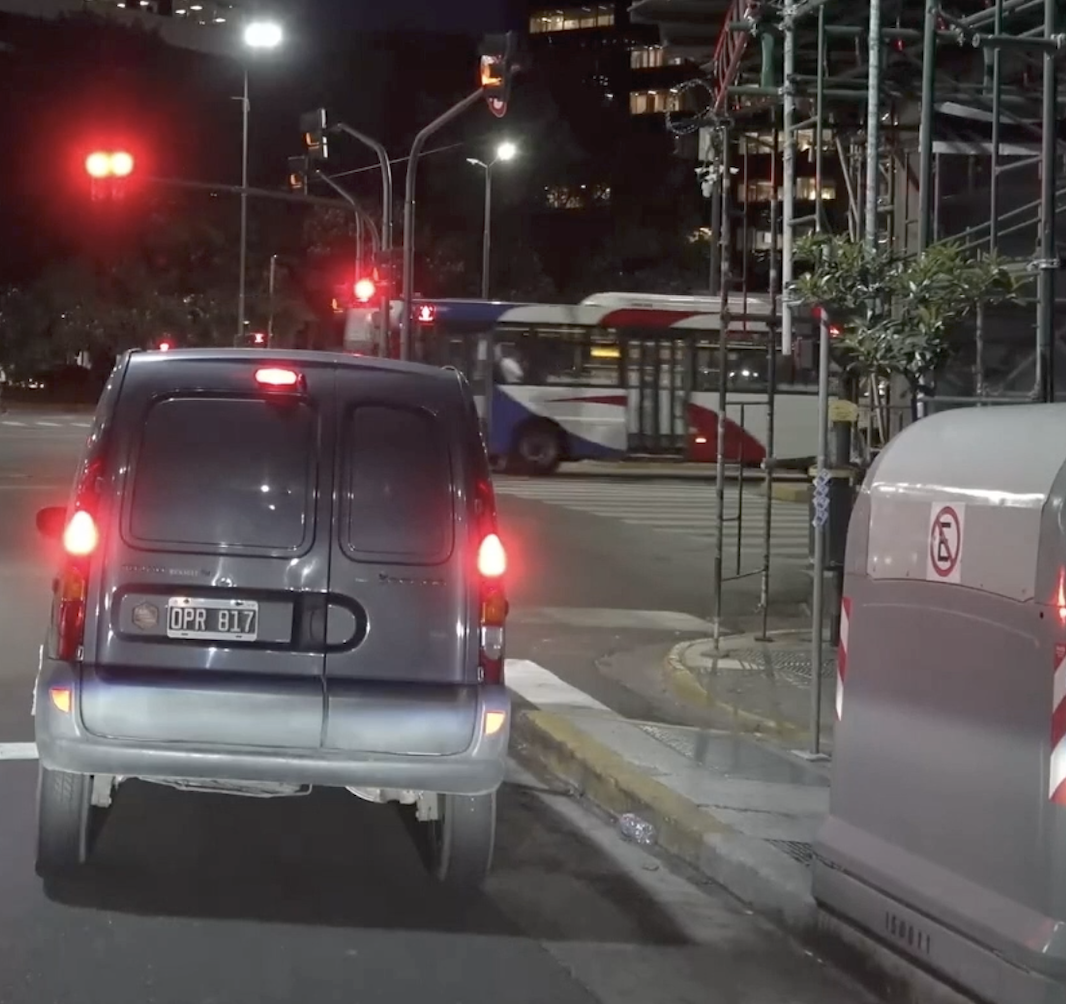}
    \end{minipage}
    \hfill
    \begin{minipage}[t]{0.315\linewidth}
        \centering
        \includegraphics[width=\linewidth]{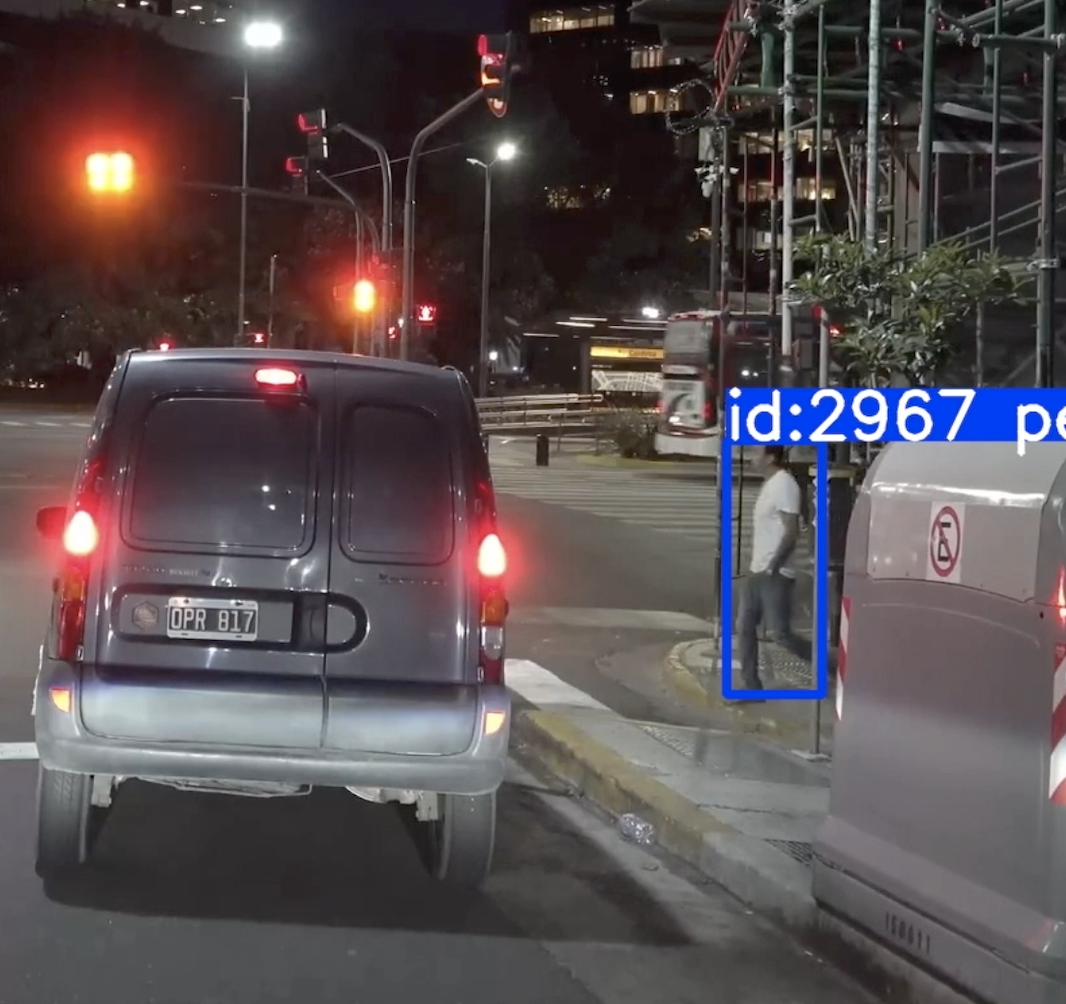}
    \end{minipage}
    \hfill
    \begin{minipage}[t]{0.315\linewidth}
        \centering
        \includegraphics[width=\linewidth]{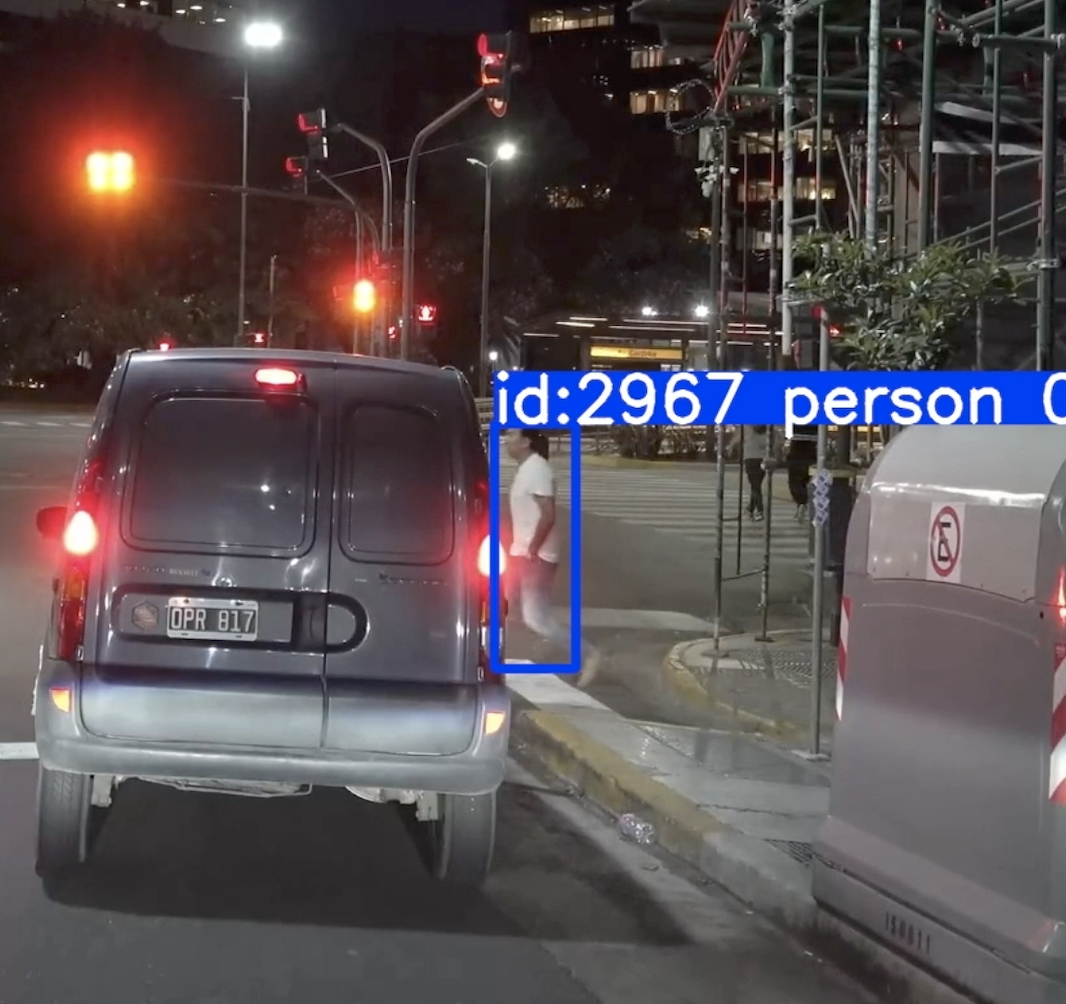}
    \end{minipage}

    \caption{Failure case with pedestrian at speed emerging behind object}
    \label{fig:failure_speed}
\end{figure}

A case from Madrid where all the models collectively make an incorrect prediction involves a pedestrian with 'track\_id' 601 waiting at the side of the road for a traffic light. It appears that the person is not crossing due to the pedestrian’s static posture, despite the pedestrian on the left of the image crossing at a red light. However, the person on the right does cross because the traffic light turns green. As the AI models and human participants do not have this information, the prediction is made solely on the basis of the pedestrian’s body posture. This case demonstrates that the prediction depends not only on the pedestrian’s behaviour, but also on other conditions that the footage does not capture. The case is illustrated in the images below.

\begin{figure}[H]
    \centering

    \includegraphics[
        width=0.95\linewidth
    ]{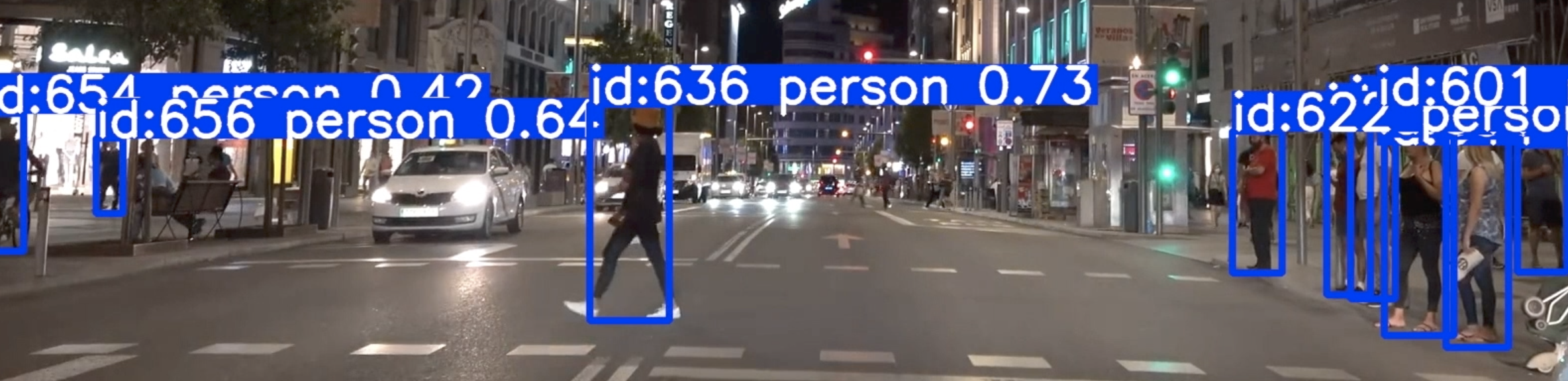}

    \vspace{0.12cm}

    \includegraphics[
        width=0.95\linewidth
    ]{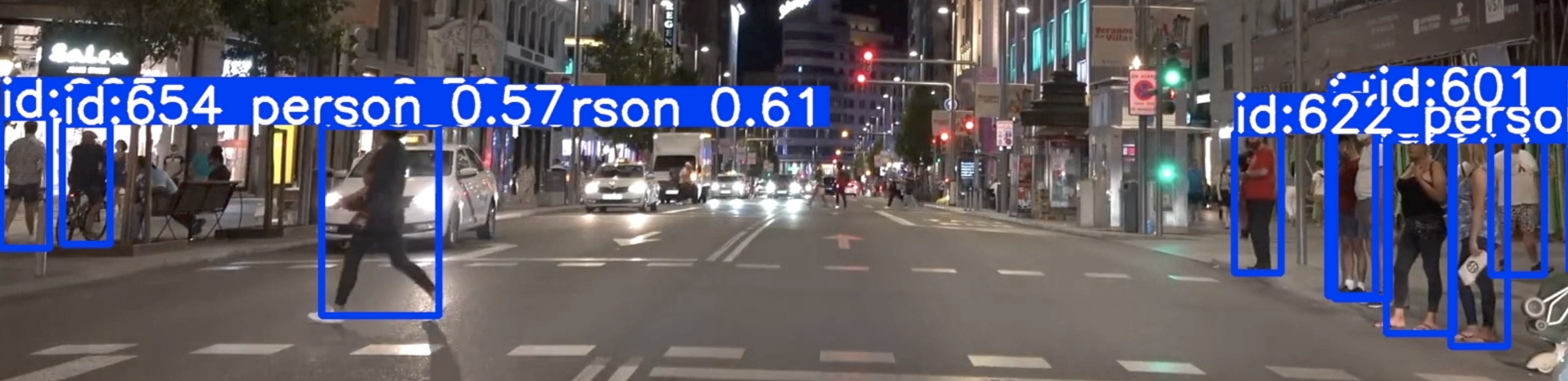}

    \vspace{0.12cm}

    \includegraphics[
        width=0.95\linewidth
    ]{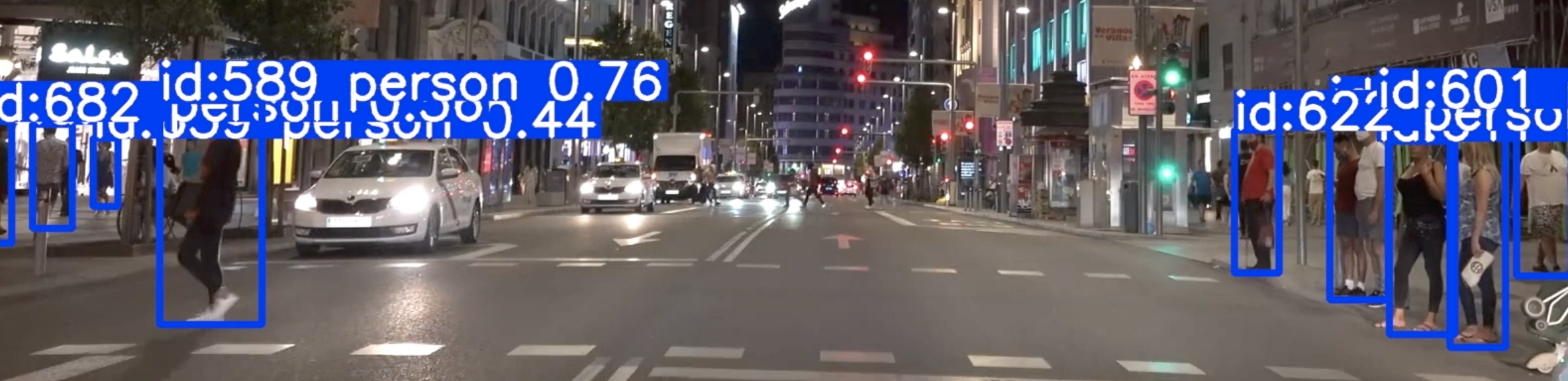}

    \caption{Failure case due to missing traffic-light information}
    \label{fig:failure_traffic_light}
\end{figure}

The final case features a clip from Tokyo showing a pedestrian with 'track\_id' 565 standing at the side of the road. Both the AI models and the human assessment make a correct prediction. In the video, the pedestrian is detected from a distance, giving the models a full two seconds to make a prediction. The pedestrian is poorly illuminated, but due to the contrast with the bright shop window in the background, the pedestrian is clearly visible. This case therefore demonstrates that the contrast between the pedestrian and the background also plays a role in predicting crossing behaviour. The images from the case are shown below.

\begin{figure}[H]
    \centering

    \begin{minipage}[t]{0.315\linewidth}
        \centering
        \includegraphics[width=\linewidth]{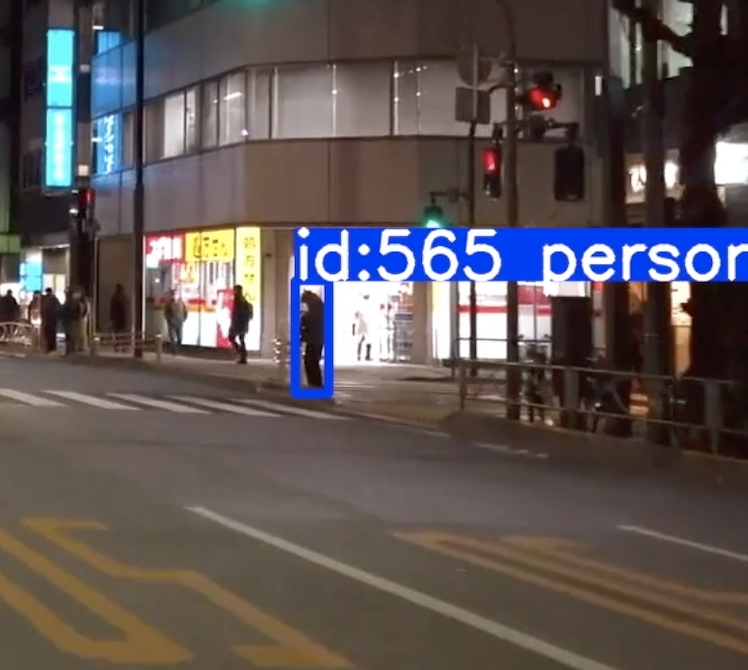}
    \end{minipage}
    \hfill
    \begin{minipage}[t]{0.315\linewidth}
        \centering
        \includegraphics[width=\linewidth]{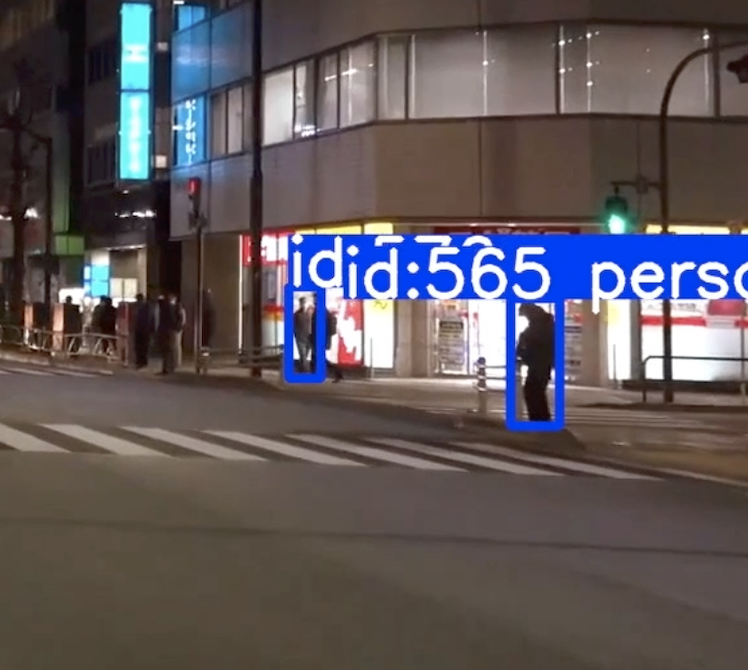}
    \end{minipage}
    \hfill
    \begin{minipage}[t]{0.315\linewidth}
        \centering
        \includegraphics[width=\linewidth]{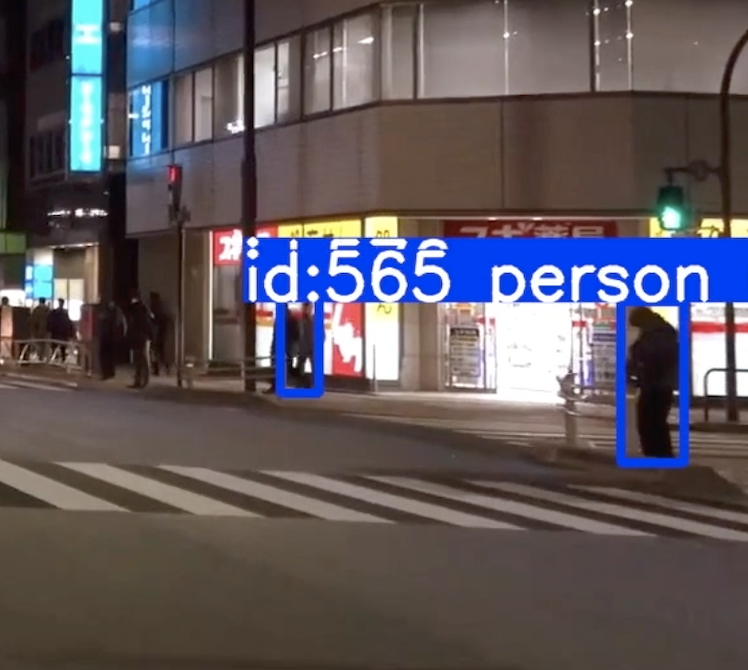}
    \end{minipage}

    \caption{Successful case with high contrast due to bright shop window}
    \label{fig:success_high_contrast}
\end{figure}

Overall, the various cases show that prediction errors arise when there is low exposure, pedestrians are grouped, partial occlusion by objects, limited contextual information, and the late appearance of pedestrians within the frame.

\section{Discussion}

Main findings show that human assessors outperform both Gemini and ChatGPT models on the benchmark. The NightSteps dataset contains a sufficient amount of challenging footage from various situations in different cities to test the performance of PCIP models. Cases involving low light, obstructions, or grouped pedestrians are considered difficult. Creating a more diverse dataset with a larger amount of visual data could lead to an even better baseline benchmark dataset in future research.

\subsection{Main Findings}

First, the various tests show that the NightSteps dataset contains sufficiently challenging visual material for models to make predictions. The cases examined demonstrate that different models do not always make the same predictions. In difficult situations where ChatGPT makes correct predictions, it shows that the model can indeed perform well, whereas a model such as Gemini fails. Other situations, however, demonstrate the opposite, with ChatGPT failing to make a good prediction while Gemini does. This shows that the strengths and weaknesses of each model can be properly evaluated and a model’s performance can be demonstrated using the generated dataset, thereby enabling specific PCIP models to be tested as well.

The results show a difference in the models’ performance across different cities. It is not only a person’s body posture that matters when predicting a crossing, but also the city’s infrastructure. In Hong Kong, for example, there is more artificial lighting visible in the video than in Amsterdam. Furthermore, traffic lights and pedestrian crossings are not always visible in the videos. This additional information can also be factored into predictions. In the video from Amsterdam, people are more likely to cross at locations where there is no pedestrian crossing than in Tokyo, where the number of crossings at pedestrian crossings is higher. 

It also appears that human predictions are more accurate than those made by the tested AI models Gemini and ChatGPT. Figures show that human predictions have an accuracy of 88.17\% and an F1-score of 80.21\%, whilst ChatGPT has an accuracy of 79.13\% and an F1-score of 63.64\%. Gemini’s performance falls even further behind, with an accuracy of 60\% and an F1-score of 42.50\%.

The failure cases show that predictions become more difficult when pedestrians are difficult to detect, are obstructed by objects, or appear in groups. It also emerges that some cases are so difficult that all models make an incorrect prediction. In addition, certain situations demonstrate that human participants themselves make different predictions. This indicates that the dataset contains various complexities, allowing for evaluation of both model performances and differences in human predictions. The variation across the results demonstrates the suitability of the dataset as a benchmark for PCIP.

\subsection{Limitations}

Although the generated dataset can serve as a solid basis for evaluating PCIP models under night-time conditions, it does have a number of limitations. These shortcomings are highlighted to ensure they are considered during model testing.

First, the tracks generated by YOLOv8 are sometimes inaccurate because pedestrians are not in all cases detected clearly in dark environments. As a result, a pedestrian may be detected and assigned different 'track\_ids' due to a lack of continuity in the detection. This means that interruptions occur in the detection, resulting in the same person being assigned a different 'track\_id' upon a new detection, even though it is still the same person. Furthermore, incomplete detection of a pedestrian also leads to a higher likelihood of an incorrect 'track\_id' being assigned. These incomplete detections result in a lower number of usable labels.

This shortage of usable clips, combined with a dataset consisting of six clips, leaves a number of 244 crossings to be labelled. This data shortage can lead to model overfitting, whereby a model learns too much from specific features in the dataset, causing it to perform less well in new environments. However, this data is intended for evaluating an already trained model, rather than training the model. A larger and more diverse dataset can counteract this overfitting.

In addition, the data contains only information about the position of the bounding boxes and the corresponding crossing labels. For many models, this information is sufficient to make an accurate prediction. However, there are a number of models like the PVIM model that, in addition to this information, also require data on a vehicle’s speed and the pedestrian’s body posture (pose). These variables can contribute to higher accuracies for PCIP. Current models should now make use of pose estimation models such as OpenPose, in which body joints and keypoints are detected to reconstruct a person’s posture and movement \parencite{cao2019openpose}. Integrating these pose prediction models is, however, feasible and makes a model more versatile.

Finally, there may be errors in the labelling of crossing behaviour. The ground-truth labels were assigned by just one person, which increases the likelihood of errors. To rule out these errors, it is essential to have the labelling process carried out by several people. This will result in a more reliable dataset, enabling models to be evaluated in an even more reliable manner.

\subsection{Future Research}

Future research can build on the benchmark dataset by expanding its size and by creating more diverse datasets. This diversity can be achieved by adding data where footage has been captured by night-vision cameras. Benchmarking models using data obtained from night-vision cameras provides information on the suitability of night-vision cameras for use once they achieve significantly higher scores in pedestrian detection and prediction. This diversity can also be created by adding  more footage from different cities. Diversity leads to a further evaluation of the model, providing greater insight into the robustness of a model.

In addition, further research could focus on testing newer YOLO models on the dataset. This study has shown that YOLOv8 is largely suitable for detecting pedestrians, but newer, further-developed models may be able to achieve even more accurate detections, even when pedestrians are small in the frame or partially obscured by objects. A comparison between newer YOLO models and the YOLOv8 model could provide insight into the models’ performance on the video footage and the usability of the dataset.

Future research could also focus on analysing pedestrian behaviour across different cities based on psychological characteristics. The period immediately before crossing is crucial for this research, allowing for an analysis of the physical characteristics exhibited by a pedestrian. These include walking speed, spatial awareness, interaction with other road users and body posture. Insight into human behaviour can be valuable as additional information in the prediction process of a model used by an autonomous vehicle, enabling even more accurate predictions to be made to improve road safety.

Finally, future research could focus on a longer observation period for pedestrians. In this study, an observation period of two seconds is used, which in some cases is insufficient to make an accurate prediction. This applies to both predictions made by the AI models and human assessments. Longer observation periods lead to the investigation of how much temporal information is required to make an accurate prediction.

\section{Conclusion}

The aim of this paper is to investigate how a dataset can be constructed and evaluated for pedestrian crossing intention prediction. To this end, the NightSteps dataset was introduced and evaluated using AI models and human assessment. This evaluation demonstrated the dataset’s suitability for testing the performance of both Gemini and ChatGPT, as well as human assessment. The main question has been answered through a number of sub-questions.

The first sub-question focused on the challenges posed by pedestrian crossing intention prediction in low-light environments. Few studies had previously investigated how PCIP performs in low-light environments due to the lack of datasets comprising night-time data. In this paper, various tests revealed that low-light environments can lead to low accuracy in pedestrian detection. The second sub-question focused on how a dataset consisting entirely of night-time data could be constructed and labelled. This was achieved by collecting dashcam footage from six different cities: Amsterdam, Buenos Aires, Hong Kong, Madrid, New York City, and Tokyo. The processed footage was used by the YOLOv8 model to detect the pedestrians’ tracks. The manually added crossing labels then provide information about the pedestrians’ crossing behaviour. The third sub-question focused on the usability of the created dataset by comparing predictions from the AI models and the human assessment with the correct ground-truth labels from the dataset. Tests on this benchmark dataset show varying results, with human assessment scoring better than the AI models. 

Overall, this research demonstrates that the night-time benchmark dataset NightSteps can be established to facilitate future evaluations of pedestrian crossing intention prediction models, all with the aim of improving road safety.

\section{Ethic Statement}

As mentioned, the dataset was partially created using videos from YouTube and footage filmed in Amsterdam showing pedestrians. This data has been used solely for scientific purposes. The data is generated based on a person’s poses, with no emphasis on a person’s facial features. Its use is always in line with various guidelines, such as the ‘Legitimate Interest for Scientific Research’ in the General Data Protection Regulation \parencite[GDPR;][]{GDPR}.

This research is also in line with the Netherlands Code of Conduct for Research Integrity and will always be conducted honestly, transparently and used solely for scientific purposes \parencite{ngwi2018}. This applies to the source of the data, how the code operates and the way the data is processed. The YouTube videos may be used in accordance with the Copyright Act (Articles 15a and 15n) provided that the video is publicly available and the authors of the videos are credited \parencite{copyrightact1912}.

.
.


{
    \small
    \printbibliography
}

\end{document}